# Highly agile flat swimming robot

Florian Hartmann[1*], Mrudhula Baskaran[2], Gaetan Raynaud[2], Mehdi Benbedda[1], Karen Mulleners[2], and Herbert Shea[1*]

## Affiliations

[1] Soft Transducers Laboratory (LMTS), École Polytechnique Fédérale de Lausanne (EPFL), Neuchâtel, Switzerland.

[2] Unsteady Flow Diagnostics Laboratory, Institute of Mechanical Engineering, École Polytechnique Fédérale de Lausanne (EPFL), Lausanne, Switzerland.

*Corresponding author. Email: hartmann@is.mpg.de, herbert.shea@epfl.ch

## Abstract

Exploring bodies of water on their surface allows robots to efficiently communicate and harvest energy from the sun. On the water surface, however, robots often face highly unstructured environments, cluttered with plant matter, animals, and debris. We report a fast (5.1 cm/s translation and 195 °/s rotation), centimeter-scale swimming robot with high maneuverability and autonomous untethered operation. Locomotion is enabled by a pair of soft, millimeter-thin, undulating pectoral fins, in which traveling waves are electrically excited to generate propulsion. The robots navigate through narrow spaces, through grassy plants, and push objects weighing over 16x their body weight. Such robots can allow distributed environmental monitoring as well as continuous measurement of plant and water parameters for aqua-farming.

## Introduction

Aquatic robots enable mapping of water pollution[1], research in marine biology[2], and deep-sea exploration[3]. Harnessing biomimicry has enabled swimming robots to blend into natural environments through silent and life-like locomotion. However, the size (>30 cm), weight (>1 kg), or propulsion strategy of those robots confine them to open and uncluttered underwater environments[2-7]. Operating on the water surface allows for monitoring of both water and air quality, while simultaneously permitting communication with terrestrial or aerial infrastructure and energy harvesting from the sun. Safely interacting with – or avoiding – animals, plants, or organic and plastic debris requires smaller and lighter robots capable of agile movements. Scenarios such as swimming through a rice field, or performing inspection through narrow openings, benefit from cm-scale (1-10 cm) robots[8]. Inspired by water-striders, rays, or amphibians, scientists have built a diverse set of cm-scale swimmers that walk[9-11], glide[12-14], and row on water[15-17], or use vibrations[18-20] or undulations[21-26] for locomotion. Most of these devices cannot maneuver or are tethered to an external power supply, limiting their applicability. Untethered operation of cm-scale robots requires mm-to-cm scale actuators, which must provide sufficient thrust for propulsion when driven by a lightweight onboard power supply. Actuation based on electro-osmotic hydrogels[15] or miniature DC-motors[11,20] are possible avenues for maneuverable, untethered swimmers, but they show moderate agility, use fragile rudders, or are bulky, and none offer autonomous decision making. Here we introduce a family of cm-scale (25 mm to 45 mm length), fast (12 cm/s tethered and 5.1 cm/s untethered), and maneuverable (120 °/s tethered, 195 °/s untethered) soft swimming robots with autonomous operation, in both energy and trajectory. Propulsion comes from traveling waves that are excited along a pair of undulating pectoral fins on the water surface, similar to polyclads – marine flatworms. The robot consists of a flat sub-mm-thin locomotion module (actuators and undulating fins) and a printed circuit board (PCB) with the power supply, control, optical sensing, and communication electronics (Fig. 1A). The locomotion modules monolithically integrated a pair of soft undulating fins with soft electrohydraulic actuators that operate at voltages below 500 V, at low power (< 35 mW), at high frequency (>100 Hz), and are durable (over 750,000 actuation cycles). Modular design strategies extend locomotion capabilities beyond forward swimming and turning to include backward and sideways swimming, offering versatile operation similar to quadcopters in air. Thanks to integrated sensors (for both infrared and visible light), energy supply, power conversion, and control, we demonstrate autonomous and untethered operation on the water surface. The swimming robots circumnavigate obstacles, swim through narrow spaces, and push away objects with 16x their own body weight. They detect moving or stationary light sources and navigate towards or away from them. This agile robotic swimmer combines advances in propulsion performance, robustness, and functionality, and will help to inspire future miniaturized vehicles operating in natural aquatic environments.

## Results and Discussion

### Locomotion design

We aimed for a robot architecture that fulfills four requirements: (1) Versatile locomotion capabilities such as directional swimming and turning, (2) facile integration of PCBs to allow untethered operation, (3) locomotion design with reduced complexity, using the minimal number of actuators, and (4) small footprint to allow navigation through narrow spaces (~5 cm) in natural environments. We addressed these requirements by developing a robot that is based on a flat locomotion module (~500-μm-thick), inspired by marine flatworms like *Pseudocerotidae*, featuring a pair of undulating pectoral fins on its sides. For flatworms, the fins extend longitudinally along the entire body and generate multiple traveling waves along the fin, aiding in locomotion and stability[27]. Similar to its natural counterpart, we realized a locomotion mode with over 1.5 wavelengths along the fin (Supplementary Fig. 1), larger than in most other artificial, small-scale surface swimmers based on undulatory pectoral fins.

The locomotion module floats on the water surface due to surface tension, benefiting from a high surface area, and can carry larger weights when combined with buoyant elements. Undulation of the fins generates flows on the water surface and beneath, resulting in propulsion. Through the individual control of each fin, we achieve high maneuverability, enabling forward locomotion and turns using only two actuators (Fig. 1B,C). Each fin is driven by a single, monolithically integrated, soft electrohydraulic actuator, powered and controlled by the onboard electronics. We designed a miniaturized multichannel high voltage power supply (HVPS), supplying two bipolar 600 V signals from a 35 x 20 mm² PCB. Downlink-communication through infrared-sensors or ambient light phototransistors, computing, and energy storage complete the circuit, allowing autonomous and untethered operation.

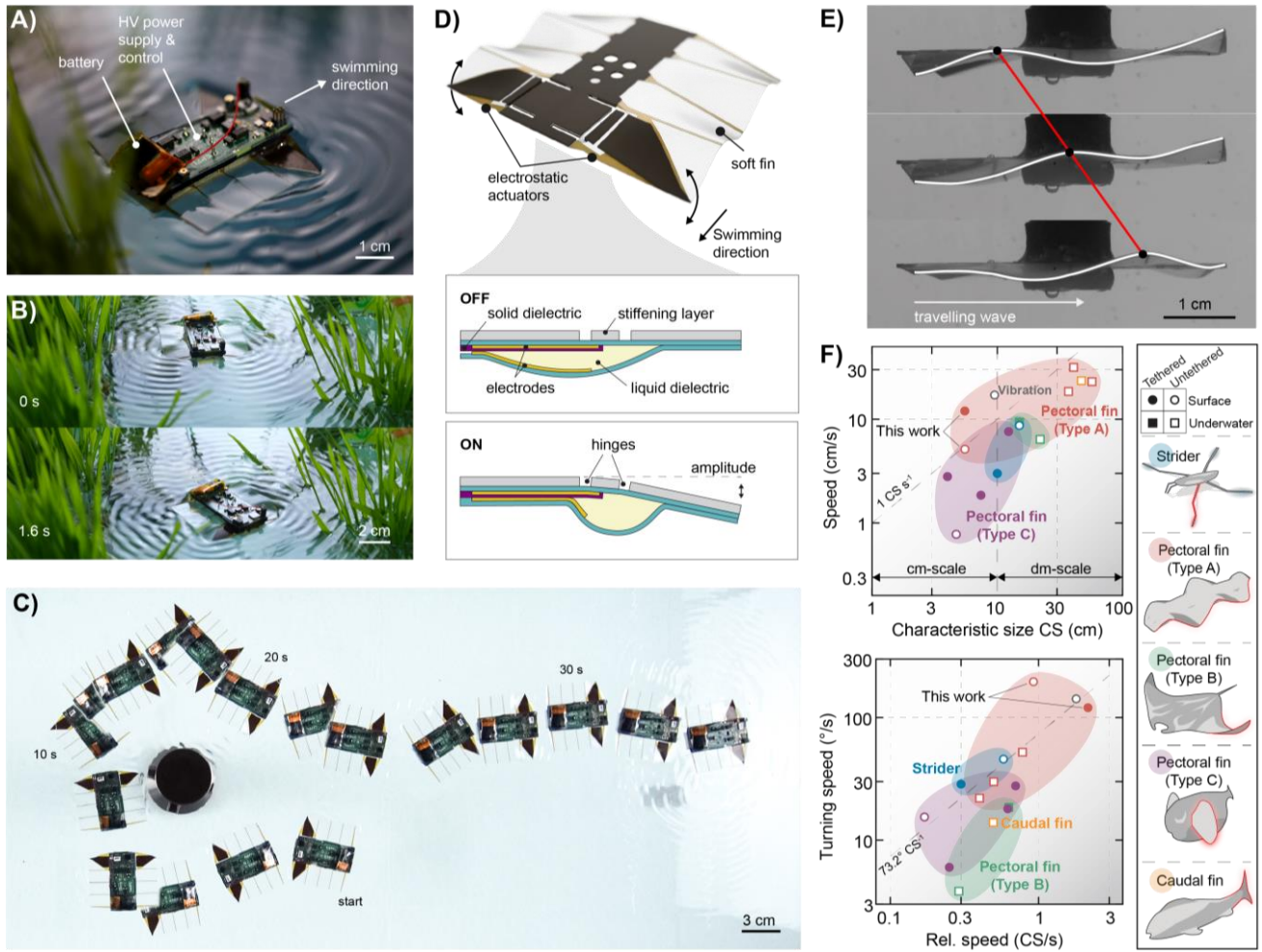

**Fig. 1 | Untethered highly agile flat robots evolving in aquatic environments.** **(A)** An untethered and autonomous robot (length, 45 mm) including miniaturized high voltage power supply and onboard control electronics. Scale bar, 1 cm. **(B)** Robot swimming on the water surface in a cluttered environment. Scale bar, 2 cm. **(C)** Individual control of the actuators allows controlled navigation of the robot, enabling turns to the left or right or to swim forward. Scale bar, 3 cm. **(D)** Schematic of a flat locomotion module using two electrohydraulic actuators to drive a pair of undulating soft pectoral fins. **(E)** Applying a periodic high voltage signal leads to oscillatory bending motion of electrohydraulic actuators and hence the generation of a traveling wave along the soft fins, both underwater and on the water surface. Scale bar, 1 cm. **(F)** Performance comparison of maneuverable swimming robots. Our cm-scale robots show high agility and speed relative to the robot's characteristic size (CS). Shaded areas mark bio-inspired locomotion modes for underwater, water surface, tethered and untethered swimmers. Dashed line in the upper plot marks a relative speed of 1 CS/s. Dashed line in the lower plot marks a linear fit of the represented data, with a slope of 73.2 °/s. Points above this line have more efficient rotation compared to the average robot. Details are given in Supplementary Table 1.

We chose soft electrohydraulic actuators due to their high performance, direct electric controllability, scalability, and extremely low acoustic noise[4,28]. Soft electrohydraulic actuators are a class of capacitive transducers that convert a high voltage input into motion. The capacitive architecture comprises two electrodes, which are separated by layers of solid and liquid dielectrics. When an electric field is applied, the electrodes zip together due to Maxwell stress and displace the liquid dielectric, which is used for deformation of the actuator (Fig. 1D). The force output depends primarily on the applied voltage and on the thickness and permittivity of the dielectric layers (see Supplementary Discussion). We designed a miniaturized and high-performance version of a type of previously reported bending actuator[29] to generate a traveling wave motion in the soft fins. In this design, a stiffening layer introduces an asymmetry in the actuator contraction, resulting in a bending motion (Supplementary Movie 1). We achieve thin mm-scale actuators (sub-mm-thickness and electrode sizes as low as 27-90 mm²) that operate at voltages (< 500 V) far below those typically used for this class of actuator, and operate at over 100 Hz. These advances are obtained through design, fabrication, and materials choices, including the use of a low viscosity fluorinated fluid as the liquid dielectric and a ferrorelaxor terpolymer as the solid dielectric.

The actuator's downward stroke is electrically driven, while the upward motion is due to the release of stored elastic energy. Periodic actuation generates a traveling wave in the soft undulating fin (Fig. 1E). We observed traveling waves for both actuation under water and on the water surface, demonstrating the versatility of this propulsion mechanism (Supplementary Movie 2). Characterizing the traveling wave underwater shows that the wavelength is inversely proportional to the actuation frequency (Supplementary Fig. 1). The combination of robot architecture, locomotion design, and high-performance miniaturized actuators, leads to high speed and controlled agile maneuverability. Our tethered and untethered designs achieve both high relative speeds (normalized to a robot's largest dimension) and higher turning speeds compared to other maneuverable bio-inspired swimmers (Fig. 1F, Supplementary Table 1).

**Performance characterization of tethered modules**

The entire locomotion module is monolithically fabricated using a combination of laser processing, lamination, and blade casting (Supplementary Fig. 2-3), similar to previously reported methods and materials[30]. The module consists of several layers of polyethylene terephthalate (PET), various adhesives, and silicone elastomer (Fig. 2A). For the actuators, we employ two different solid dielectrics: (1) a 12-μm-thick PET film with relative permittivity of 2.2, or (2) a 14-μm-thick layer of a ferrorelaxor terpolymer (poly(vinylidene-fluoride–trifluoroethylene–chlorotrifluoroethylene), P(VDF-TrFE-CTFE), here referred to as PVDF-terpolymer) with high relative permittivity (~40). We utilize PET as dielectric due to an easier fabrication process and to explore design rules for the robots, while the PVDF-terpolymer is used to lower the operating voltage of the actuators to below 500 V to enable untethered operation. We fill the modules with a fluorocarbon liquid (FC 40) due to its low viscosity, which allows high frequency actuation. A typical module, with a size of 45 x 55 x 0.5 mm³ weighs 1.23 g, which is light enough to float on the water surface by surface tension.

We first tested the performance of tethered modules in free-swimming tests, for which we connect the module using thin copper wires (50 μm diameter) in combination with a magnetic connector (Fig. 2B). The modules are driven with a bipolar rectangular voltage signal (typical amplitude is 1700 V, Supplementary Fig. 4) to avoid interfacial charging at the interface of solid and liquid dielectric[31]. We observed that the swimming speed depends on the actuation frequency and is highest at 40 Hz, where modules achieve on average a speed of 7.5 cm/s ± 0.9 cm/s (1.7 BL/s ± 0.2 BL/s), indicating benefits from resonant behavior (Fig. 2C).

In tethered modules, the bending stiffness of the wires constrains the module's displacement in free-swimming tests, preventing prolonged swimming speed or lifetime measurements. We therefore measured the blocked propulsion force, thus avoiding the mechanical influence of the wires. In this configuration, the module is made to swim against a glass cantilever (Fig. 2D, Supplementary Fig. 5, Supplementary Movie 3). The blocked force is calculated from the cantilever deflection (see Supplementary Discussion). This measurement method is suitable to optimize the driving conditions, varying design parameters, and measuring lifetime for a large number of swimming modules. For a drive voltage of 1500 V, we obtained the highest blocked force of 1.1 mN (1.6 mN at 1700 V, Supplementary Fig. 6) at 40 Hz actuation frequency, the same frequency where the maximum swimming speed was measured (Fig. 2E). When using the PVDF-terpolymer as dielectric, a blocked force of 0.8 mN is reached at 30 Hz actuation frequency at 500 V. This frequency-shift when changing dielectrics stems from differences in the thickness of the encapsulation layer in both variants (Supplementary Fig. 2). Using the high permittivity dielectric, the operating voltages are significantly reduced: only about a third of the operating voltage is required when switching from PET to the PVDF-terpolymer, in agreement with our analytic models (Fig. 2F, Supplementary Discussion). The blocked force as a function of applied voltage shows a linear dependence on voltage, starting from threshold voltages of about 400 V and 1200 V using the PVDF-terpolymer or PET, respectively. Typically, both dielectric variants show runtimes of over 2.5 hours, corresponding to over 270k actuation cycles (Fig. 2G) and over 750k cycles in some cases (Supplementary Fig. 7A). The initial performance stays stable for a few minutes before dropping to 40-50%, where modules with the PVDF-terpolymer usually show a faster decrease. This reduction in performance mainly stems from mechanical fatigue of the actuator pouches, while electrical breakdowns play a minor role (Supplementary Fig. 7B). Despite the use of thin metal electrodes and moderate voltages, electrical breakdowns lead to ablation of the electrodes in the region of failure, without damaging the shell material. This self-clearing effect lets the actuators withstand several breakdown events with minimal effect on the actuator's performance.

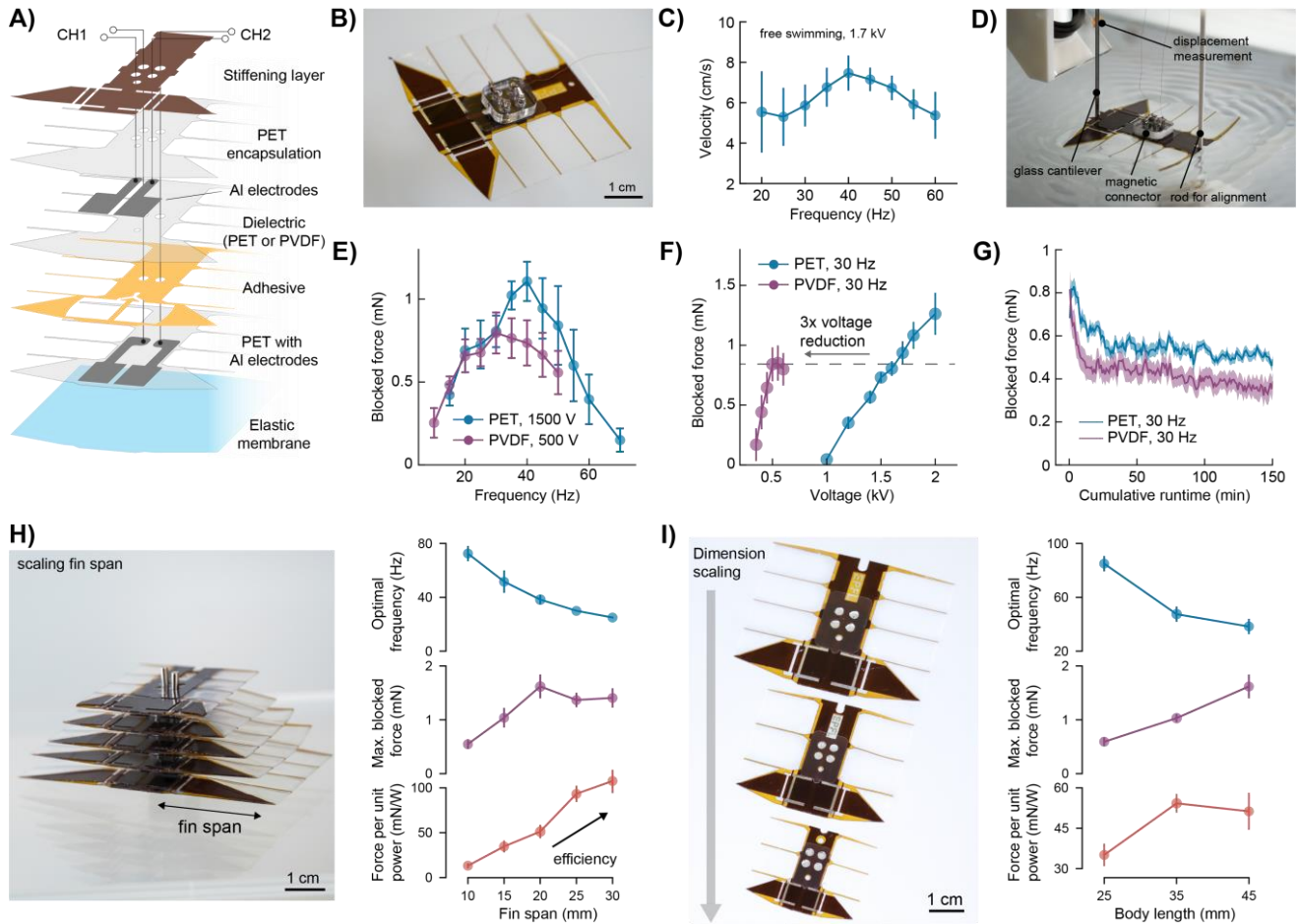

**Fig. 2 | Performance and scaling analysis of tethered soft locomotion modules.** **(A)** The locomotion module consists of multiple polymer layers and metal electrodes, enclosing an oil-filled pouch. **(B)** The module locomotes on the water surface by generating traveling waves. It is connected to an external power supply through thin copper wires and a magnetic connection system, adding 2.25 g to the module. Scale bar, 1 cm. **(C)** Free-swimming tests of modules with PET dielectric show the highest speed at 40 Hz actuation frequency. Applied voltage is 1.7 kV. **(D)** A measurement of the blocked propulsion force allows controlled characterization of the swimming performance for extended durations. **(E)** The blocked force is maximum at an actuation frequency of 40 Hz and 30 Hz for modules using PET and PVDF-terpolymer as dielectric, respectively. **(F)** The blocked force at a fixed frequency of 30 Hz shows a nearly linear trend with applied voltage. Using the PVDF-terpolymer as dielectric layer enables a 3-fold reduction of the operating voltage. **(G)** A cumulative runtime of more than 2.5 h is achieved for both dielectrics. With the applied duty cycle of 1:5, the modules were operational for over 15 h. **(H)** Modules with scaled fin span but equally dimensioned actuators and body length. Modules with larger fin span operate best at lower actuation frequencies and deliver a high blocked force and, hence, show increased swimming efficiency. Scale bar, 1 cm. **(I)** Modules with constant aspect ratio and scaled body length of 45, 35, and 25 mm. Larger modules have lower optimal actuation frequencies while achieving a higher maximal blocked force. The swimming efficiency is higher for larger modules. Scale bar, 1 cm.

Using the measured blocked propulsion force as a performance metric enabled us to compare various robot designs (a design overview is given in Supplementary Fig. 8). We performed a scaling analysis of two major design parameters: (1) scaling the fin span (i.e., width) of the module at constant body length, and (2) scaling of the module's footprint (i.e., width and length) at constant aspect ratio. For the first scenario, we keep both the body length (45 mm) and the size of actuators constant, while scaling the span of each fin from 10 mm to 30 mm (Fig. 2H). This scaling has a threefold effect on swimming performance: First, we observed that the optimal actuation frequency, which is 40 Hz for the module with 20 mm fin span, decreases with larger fin span, due to the changing resonant behavior. Second, the corresponding (maximal) blocked force increases with larger fin span but reaches a limit for 20 mm and larger fins. This indicates that a maximum energy conversion from electrical input to propulsion is reached. Third, the ratio of blocked force to input electrical power, a quantity that relates to swimming efficiency, increases with higher spans. The power consumption of soft electrohydraulic actuators is directly proportional to the actuation frequency and the capacitance of the actuators in their fully actuated state (Supplementary Fig. 9). The largest module with a fin span of 30 mm, for example, achieves its highest blocked force (1.4 mN) at 25 Hz while requiring 13 mW for actuation. Similar trends are observed when uniformly scaling the robot's width and length. We investigated three module sizes measuring 45, 35, and 25 mm length and observed that optimal actuation frequencies increase for smaller modules, while the blocked force decreases (Fig. 2I, Supplementary Fig. 6). The force per unit power increases with size, indicating more efficient swimming for larger modules. The indifference between 35 mm and 45 mm sized modules stems from slight variations in the used encapsulation layers, as a thicker encapsulation was used in the 45 mm module. As a compromise between performance, footprint, and energy efficiency, modules with 45 mm and 20 mm fin span represent the preferred choice for untethered robots.

**Swimming mechanisms and maneuverability**

Our robot swims and steers using a pair of undulating pectoral fins (Fig. 3A). While this propulsion mechanism is commonly found for animals underwater, it is rarely used –if at all– by animals on the water surface. The relation between actuation frequency and swimming speeds resulted in finding optimal driving parameters. Here, we visualize the generated surface flow, using line integral convolution (LIC) and particle image velocimetry (PIV) to get insights on the mechanisms of propulsion (Fig. 3B,C, Supplementary Movie 4). For forward propulsion, both fins push flow to the side and the back to generate counter-rotating recirculation regions behind the robot, which produce a thrust wake pushing the module forward (Fig. 3B). Similar patterns that generate thrust are observed for water striders and honey bees, swimming at the air water-interface[32,33].

The velocity of the backwards flow (relative to the module) and the width of the recirculation region have similar frequency behavior than the module's performance. They show an optimum at 40 Hz, correlating performance to the generation of surface flows (Supplementary Fig. 10). If only a single fin is driven, the recirculation region becomes asymmetric and results in the rotation of the robot (Fig. 3C). In addition to the surface flows, the robot generates flows underwater, which we qualitatively visualize in Supplementary Movie 5. These results indicate that propulsion is caused by a combination of surface and underwater flows, similar to the hydrodynamics observed in a recent study[26], and investigating the ratio of those contributions will be part of future studies.

The combined effects of propulsion enables the modules to reach high speeds. For forward propulsion, the fastest module accelerates to 12 cm/s (2.6 BL/s) within a few seconds, showing a peak acceleration of 38 cm/s² at 2 kV (Fig. 3D). In some cases, we observed that the locomotion module deviated from a perfectly straight path, resulting from small differences in the fabrication of the actuators. In the future, tuning the actuation frequency of each fin individually using a closed loop control, will allow swimming along complex trajectories supported by path planning. The undulating fin design and fast acceleration lead to high agility. Locomotion modules turn almost on the spot (turning radius as small as 20 mm) with an angular speed of about 120 °/s at 1.7 kV (Fig. 3E, Supplementary Movie 6).

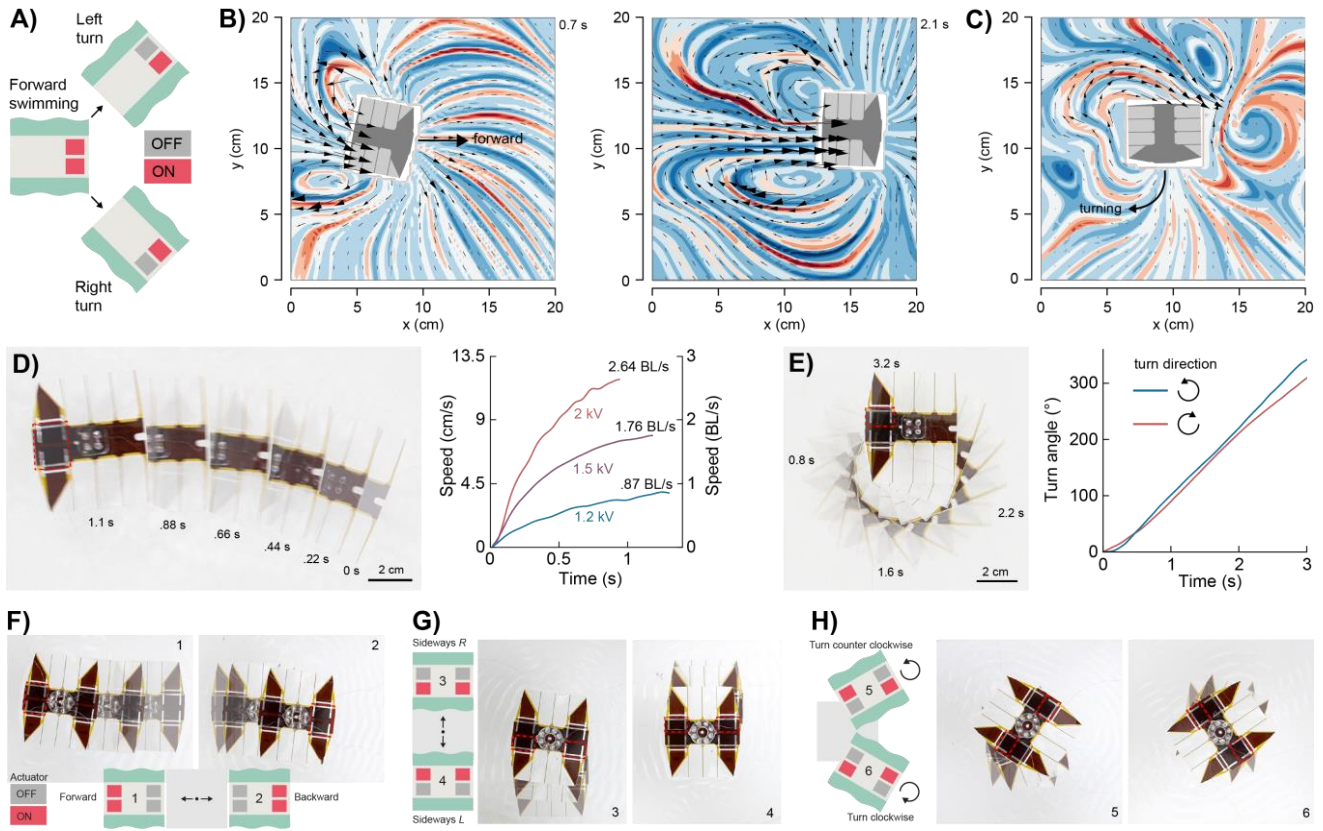

**Fig. 3 | Versatile maneuverability through 2- and 4- actuator designs. (A)** Control scheme for a 2-actuator locomotion module. Forward swimming is achieved by applying voltage to both actuators, while rotation is achieved by using a single actuator. **(B, C)** Flow visualizations using Line Integral Convolution show **(B)** two counterrotating recirculation regions symmetrically located behind the locomotion module that yield forward propulsion, and **(C)** an asymmetric wake pattern when the robot is turning. Black arrows show the surface velocity field. **(D, E)** Swimming and rotation speed for the best performing tethered locomotion module. **(D)** The locomotion module reaches up to 2.6 BL/s (12 cm/s) for forward motion and **(E)** 120 °/s for turning for both directions. Scale bar, 2 cm. **(F-H)**, A 4-actuator design increases functionality by adding backwards (2) and sideways (3, 4) motion to forward (1) and turning (5, 6) motions. The module can seamlessly switch between **(F)** forward (6.7 cm/s) and backward (6.4 cm/s) motion through using the front or rear pair of actuators. **(G)** Sideways locomotion is enabled by using either both left or right actuators, resulting in directional locomotion at reduced speed (2.8 and 4.1 cm/s). **(H)** Rotations are achieved by either using a single actuator or a diagonal pair of actuators. Scale bar, 2 cm.

The locomotion capabilities are further enhanced by adding two additional actuators on the rear side of the robot, mirroring the module at its central axis. Depending on the active pair of actuators, the module now swims forward, backward, sideways, or turns, rendering a vehicle similar to quadcopter drones but for the water surface (Fig. 3F-H, Supplementary Movie 7). Notably, this 4-actuator-module reaches similar speeds for both forward (6.7 cm/s at 1.7 kV) and backward (6.4 cm/s) locomotion, which is comparable to 2-actuator-modules (Supplementary Fig. 11). Sideways locomotion is achieved when two actuators of the same side are activated, resulting in speeds up to 4 cm/s. Turning is achieved by either using a single actuator or a pair of actuators along the diagonal. In the latter case, the module turns with angular velocities up to 95°/s. These results demonstrate that with a limited amount of actuators, a full range of maneuvers is achieved, allowing sophisticated operation scenarios and navigation through complex terrains.

**Autonomous untethered swimming robots**

Operation in most natural environments requires the robot to be untethered. To this end, we developed a small PCB that includes all elements needed to operate autonomously (Fig. 4A). A key element on the PCB is a multichannel HVPS that provides two individually controlled channels of bipolar voltage signals at over 600 V, driving the two actuators using the PVDF-terpolymer as solid dielectric (Fig. 4B). In addition to the HVPS, the PCB hosts a microcontroller (MCU) and sensors for remote control and autonomous operation (Supplementary Fig. 12). To generate a high voltage output, the HVPS amplifies the input from a lithium-ion battery (~3.7 V) to over 500 V using an open-loop flyback converter circuitry (FBC)[34]. The magnitude of the output voltage is mainly influenced by the pulse width and frequency of the MCU's control signal (Fig. 4C). Aiming for an output voltage higher than 500 V, we found that a frequency of 20 kHz with a pulse width of 2.5 µs is a practical compromise between output voltage and power consumption (about 500-600 mW, see Supplementary Discussion). The high voltage is supplied to the actuators via two individually controlled full bridges, which switch the output, including its polarity, to charge and discharge the actuators. The magnitude of the voltage delivered to the actuators depends on how many actuators are active and what actuation frequency is used (Fig. 4D, Supplementary Fig. 12). For 30 Hz actuation frequency, the HVPS supplies up to 710 V to a single actuator and 620 V if both actuators are addressed. We steer the robot using a four-button remote control based on infrared communication. An IR receiver on the robot's PCB decodes the control signals and forwards them to the MCU. We integrate the PCB with the battery onto a 3d-printed tray, which protects the electronics from water, and connect it to the locomotion module. As the added weight (~5 g) of the PCB, battery, and tray surpasses the weight limit of the locomotion module (Supplementary Fig. 13), we add a thin foam layer under the locomotion module to increase buoyancy, resulting in a total weight of the robot of 6.23 g

The untethered robot swims forward with up to 5.14 cm/s (1.14 BL/s) and makes turns to both left and right, with up to 195 °/s (Supplementary Fig. 14). Given its speed, robustness, and maneuverability the robot is capable of swimming through narrow spaces and pushing away obstacles (Supplementary Movies S8-10). We show that the robot can displace obstacles weighing 16 times (101 g) its own body weight (Fig. 4E,). Once engaged with a targeted obstacle, a series of turning movements lead to detachment of the robot from the obstacle and to continuation of its task.

To go from remote control to closed loop operation, we installed four phototransistors around the robot (Fig. 4F). The robot compares the measured light intensities and is programmed to swim towards the direction of highest intensity. This added autonomy lets the robot follow a moving light source or swim towards various light sources guiding a task (Fig. 4G, Supplementary Movies S11-12). Notably, the general use of a PCB allows incorporating a wide range of sensors and communication systems that can be chosen for specific tasks. Chemical sensors for the monitoring of water quality, cameras to explore unknown environments, or hydrophone and sonar are examples of future applications.

**Conclusion**

In summary, we demonstrated a cm-scale, untethered, and autonomous swimming robot with high speed and agility. The flat, monolithically fabricated robot allows robust locomotion on the water surface and enables practical applications in cluttered environments. The frugal, yet modular use of high performance electrohydraulic actuators to enable undulatory propulsion, will pave the way for a new generation of aquatic drones that support applications in environmental monitoring, aqua farming, and agriculture.

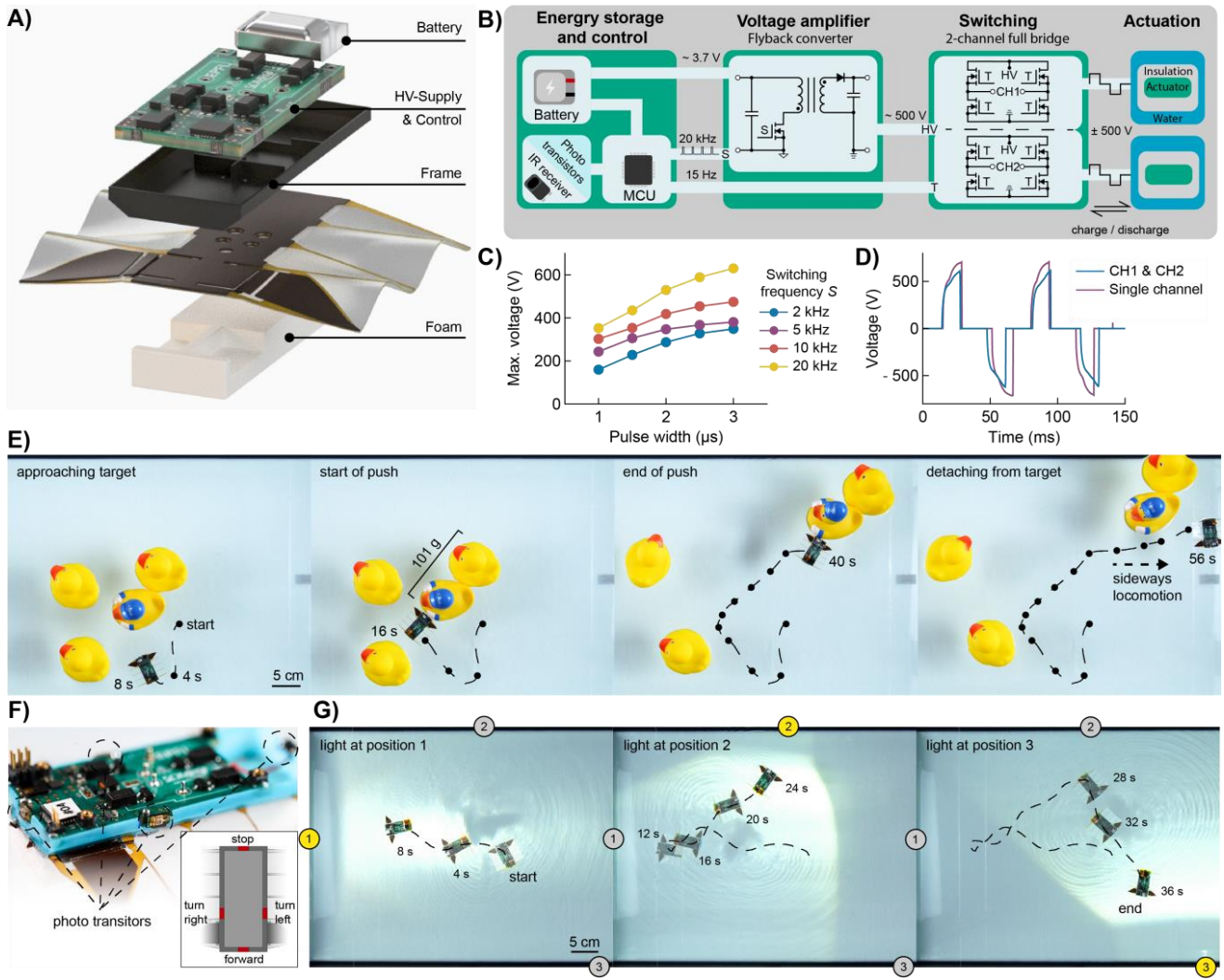

**Fig. 4 | Untethered remote controlled and autonomous swimming robots. (A)** Exploded schematic view of the untethered robot. **(B)** Simplified block diagram of the HV power supply, control components, and their connections to the actuators (see also Supplementary Fig. 12). **(C)** Maximal output voltage of the flyback converter as function of the control signal, which is supplied to switch $S$ on the primary side of the transformer. **(D)** Voltage measured at the input of the actuators as function of time and for forward control (CH1 & CH2) and rotation (single channel), using a switching frequency of 20 kHz with a pulse width of 2.5 µs. **(E)** Remote controlled robot approaching, pushing away, and detaching from obstacles that are 16x heavier than the robot itself. Points on the trajectory indicate the robot's position in 4 s intervals. Scale bar, 5 cm. **(F)** Autonomous robot equipped with four phototransistors to detect the direction of an external light source. **(G)** Autonomous robot swimming in the direction of external sources of white light, which are activated in sequence. Scale bar, 5 cm.

**Acknowledgments:** We thank Prof. Klas Hjort for discussion on the propulsion mechanism. We thank Sanjay Manoharan for assistance in 3D-printing, Javier Tavalante for discussion on MCU programming, and Jeremy La Scala for assisting early development of characterization setups. We thank Robert Hennig, Michael Smith, and Giulio Grasso for assistance and advice in taking videos and photos.

**Funding:** This project has received funding from the European Union's Horizon 2020 research and innovation programme under grant agreement No 101016411 "SOMIRO".

**Author contributions:** F.H and H.S. conceptualized the work; F.H. designed and fabricated the devices, carried out the experiments, analyzed the data, and created the figures and videos, with supervision from H.S. M.Baskaran, G.R, and F.H conducted PIV, LIC, and underwater experiments, analyzed and visualized the data, with supervision from K.M. M.Benbedda designed, programmed, and characterized the electronic circuit board, with inputs from F.H. and H.S. F.H and H.S. prepared the original draft of the manuscript; All authors reviewed and edited the manuscript.

**Competing interests:** The authors declare no competing interests.

**Data and materials availability:** All data are available in the main text or the supplementary materials.

# Supplementary Information for

## Highly agile flat swimming robot

Florian Hartmann[1*], Mrudhula Baskaran[2], Gaetan Raynaud[2], Mehdi Benbedda[1], Karen Mulleners[2], and Herbert Shea[1*]

**Affiliations**

[1] Soft Transducers Laboratory (LMTS), École Polytechnique Fédérale de Lausanne (EPFL), Neuchâtel, Switzerland.

[2] Unsteady Flow Diagnostics Laboratory, Institute of Mechanical Engineering, École Polytechnique Fédérale de Lausanne (EPFL), Lausanne, Switzerland.

*Corresponding author. Email: hartmann@is.mpg.de, herbert.shea@epfl.ch

**The Supplementary Information includes**

Materials and Methods

Supplementary Discussion

Supplementary Figs. 1 to 14

Supplementary Tables 1 to 2

Supplementary References (*35–44*)

Supplementary Movie Captions 1 to 12

## Materials and Methods

<u>Fabrication of robots with PET dielectric:</u> 12-µm-thick Mylar (a type of biaxially-oriented PET) metalized with 30 nm aluminum (ES301913, GoodFellow) was used for the electrode layer. An etching mask was printed onto the metalized film using a solid ink printer (Xerox ColorQube 8580) and the electrodes were etched using a 2% weight potassium hydroxide solution at room temperature. The etch mask was then transferred from the Mylar to regular printing paper using a laminator (Matrix Duo MD-460, Vivid) at 120°C. The electrode layers were then laser-cut (Speedy 300, Trotec). A thermal adhesive (Thermal Bonding Film 583, 3M) was cut and laminated first onto the bottom electrode layer (with the electrode facing the adhesive) at 120°C, then onto the Mylar side of the top electrode layer at 135°C, resulting in a Mylar-Al-Mylar-Al structure. We added a sealing layer onto the top electrode to prevent contact with water. For this layer, we cast a 10-µm-thick layer of thermal adhesive (Vitel 3350B from Bostik, dissolved in methyl ethyl ketone (MEK)) onto a 25-µm-thick Mylar, structured it using laser cutting, and laminated it onto the top electrodes at 120°C. For modules with 35 mm and 25 mm BL, we use a 10-µm-thick layer of polystyrene (PS, Sigma Aldrich) as encapsulation instead of PET. A stiffening layer is cut from 250-µm-thick PET (McMaster) and fixed onto the top side of the layer structure with a double-sided adhesive (50-µm-thick, ARcare, Adhesives Research). The assembly is again structured using a laser cutter, to define the soft fins. We cast a layer of Ecoflex-30 (Smooth-On) on the bottom side of the assembly and cured it at 65°C for 4h. Before the elastomer is cast, the bottom side of the assembly was activated with O2 plasma (70 W, 3 min, Diener Atto) to increase adhesion. We laser-cut the robot again to define its final shape. The dielectric liquid (Fluorinert FC-40, 3M) is injected under constant pressure of 30 mbar through a filling hole and sealed using the double-sided tape and a 25-µm-thick layer of Mylar. For all locomotion modules, we use a single port to fill multiple actuators. This port is sealed after filling but leaves a small channel that connects all actuators, even during operation. We found that this method is a frugal way to equally distribute the liquid in the pouches, maintain equal filling pressures, and obtain similar performance of all actuators on the module. A typical locomotion module with 45 mm BL has 55 ± 5 µl of dielectric liquid filled into both actuators.

<u>Fabrication of robots with PVDF dielectric:</u> For locomotion modules with PVDF-terpolymer as dielectric, the same steps are applied for fabricating the electrodes as described above. A 14 % weight solution of PVDF-TrFE-CTFE (Piezotech® RT-TS, Arkema) in MEK is cast on the top electrode layer using a thin film applicator (Zehntner ZAA 2300) with a gap of 288 µm resulting in a roughly 14-µm-thick layer of PVDF-TrFE-CTFE. During the casting process, we use a mask (GP20, Nexus, 80 µm) to pattern the dielectric around the electrodes only. The top electrode layer with PVDF dielectric is annealed for 1 h at 105°C. Bottom and top electrode layers are laminated at a maximum temperature of 120°C, resulting in a Mylar-Al-PVDF-Al-Mylar structure. As the top electrode is already isolated from water through its Mylar substrate, an additional sealing layer is omitted in this design. The fabrication further follows the steps as described above. During the assembly of bottom and top electrodes, a conductive silicone-based adhesive (SS-27, Silicone Solutions) is applied on the rear side of the locomotion module to have the contact pads of the top electrode facing upwards.

<u>Traveling wave underwater:</u> We produced a travelling wave underwater (Fig. 1e, Supplementary Movie 2) supplying a bipolar triangular high voltage signal with an amplitude of 2 kV. The actuation frequency of the 45 mm long module is 16 Hz. The sequence was filmed using a high-speed camera (Phantom VEO310L).

<u>Free swimming measurements:</u> Locomotion modules with PET dielectric were positioned on the water surface and connected with a magnetic connector (2.25 g) and two copper wires (50 µm diameters) to an external power supply (Trek 609E-6). We supplied a bipolar rectangular signal at various frequencies and voltages to drive the robot (Supplementary Fig. 3). The swimming speed is determined by filming the robot from the top and image tracking using Kinovea software.

<u>Blocked force measurements:</u> The robots were positioned in front of a hollow glass cantilever (0.05 mm x 1.00 mm, Hollow Rectangle Capillaries, CM Scientific) and on the water surface of a tank measuring 30 x 20 x 20 cm³ (halfway filled). Each robot was connected with the magnetic connector and two copper wires (50 µm diameters) to an external power supply (Trek 609E-6). We supplied a bipolar rectangular signal at various frequencies and voltages to drive the robot. The deflection of the cantilever is measured with a laser displacement sensor (IL-065, Keyence) and recalculated into a force value (see Supplementary Discussion). The robot is activated for 5 s for each driving parameter and the mean blocked force is determined between 2.5 and 5 s after the robot starts to swim.

<u>Load carrying measurements:</u> Locomotion modules with PET dielectric are placed into the setup for measuring the blocked force and sequentially loaded with evenly distributed weights. Robots are tested at 40 Hz and at 1.5 kV and 1.7 kV.

<u>Lifetime measurements:</u> The locomotion modules are placed into the setup for measuring the blocked force. We test modules with PET dielectric at 30 Hz and supplying 1500 V and modules with PVDF-terpoymer dielectric at 30 Hz and 500 V. For both designs we let the robots continuously swim for 5 min followed by a 10 min break. We repeated this cycle until the failure of the modules.

<u>Actuator power consumption:</u> The power consumption was measured using the voltage and current monitor of the Trek high voltage amplifier. Each measurement was calibrated by measuring the power consumption when no load (locomotion module) was connected to the power supply.

<u>Traveling wave metric measurements:</u> Locomotion modules were connected with a stiff, 3D-printed connector to a custom-made power supply. The modules were placed underwater and filmed from the side using a high-speed camera (Photron SA-X2 with Canon EF 100mm f/2.0 lens) at an image acquisition rate of 500 to 1000 Hz, depending on the actuation frequency of the locomotion module. A light sheet generated with a high-power light-emitting diode (ILA_5150, LED Pulsed Systems) was used to illuminate the edge of the module's fin. We supplied a bipolar rectangular voltage signal (1500 V amplitude) and recorded the fin profiles for various frequencies.

<u>LIC and PIV measurements:</u> Locomotion modules (PET dielectric) are placed on the water surface of a tank measuring 30 x 30 x 60 cm³ and filled half-way. Particles from milled black pepper are added to the water surface. A LED panel with a light diffuser is placed below the water tank to provide continuous back-illumination of the particles. A high-speed camera (Photron SA-X2 with Canon EF 100 mm f/2.0 lens) records the surface motion from above. Recording is manually started and stopped to capture the full motion of the robot, including the acceleration and deceleration phases. The modules are driven by a custom-made multichannel power supply, typically supplying a bipolar rectangular signal of 1.7 kV and at various frequencies.

Images were acquired at 250 Hz and 500 Hz by the high-speed camera with a definition of 1024 x 1024 pixels. Particle images were correlated using a multi-grid evaluation method with a final window size of 96 x 96 pixels and a step size of 10 pixels (overlap of 90%), resulting in a physical resolution of 2.04 mm.

The position and orientation of a module are obtained for each frame by finding the maximum cross-correlation of the image and the module's outline. A rectangle is used to mask the module during PIV processing. The signals in the masked regions are not correlated during PIV processing.

Line integral convolution (LIC) figures indicate the flow topology by filtering color patterns along local streamlines(35). Two convolution passes are applied to pink noise patterns with a normalized standard deviation of 0.15.

<u>Measurement of HVPS output voltage:</u> The PCB was connected to two external power supplies (Aim-TTI EL302, max. current 2 A) to find an ideal configuration resulting in high FBC output voltage and reasonable power consumption. The first one, set to 3.9 V, replaced the battery, while the second one, set to 3.3 V, powered the control circuit of the PCB. We varied the pulse width of the signal used to control the switch on the primary side of the FBC (within range 1-3 μs) and the frequency of this signal (within range 1-20 kHz). Voltage output was measured in four configuration: (1) at the output of the FBC with none of the full bridges active, (2) at the output of CH1 with only the first full bridge active, (3) at the output of CH2 with only the second full bridge active, and (4) at both CH1 and CH2 with both full bridges active. Output voltages were measured using a high voltage probe (GW Instek GDP-050) and an oscilloscope (Analog Discovery Pro ADP3450).

Measurement of power consumption: For measuring the power consumption, the FBC was either off or driven under the constant conditions of 2.5 µs pulse width at 20 kHz and 15 Hz full bridge switching frequency, which results in a 30 Hz actuation frequency. The entire PCB was powered by a Lithium-ion battery. A 1 Ohm resistance was inserted between the negative side of the battery and the PCB to measure the current and calculate the power consumption. Simultaneously, the output voltage of the FBC was measured as described above.

Untethered robot: For the assembly of untethered robots, we choose locomotion modules of 45 mm length and 55 mm width and with PVDF-terpolymer as dielectric. Plastic trays are 3D-printed (Aqua Hyperfine Resin, Phrozen) and post-processed by laser machining. First, the PCB (designed in Altium designer, produced by Eurocircuits, components are listed in Supplementary Table 2) and the tray are assembled. To this end 6.9 mm long pins (Nail Head Pin, Mill-Max), serving as connections to the locomotion module, are guided from the bottom side through both the tray and PCB and soldered from the top. Moisture curing conductive silicone (SS-27, Silicone Solutions) is applied to the connection pins to establish contact to the locomotion module. Thin lines of silicone sealant (Sil-poxy, Smooth-On) are applied on the locomotion module and the tray and PCB are placed on top of the module. Sealants and conductive silicone are cured for a minimum of 4 h at room temperature. Two layers of polystyrene foam (3-mm thick each) are cut and glued on the bottom side of the robot using UV-curable glue (Bluefixx repair). A lithium-ion battery with a nominal capacity of 30 mAh and a manual on/off switch are mounted on the PCB, finalizing the robot. For the autonomous version of the robot, we soldered four phototransistors on each side of the robot.

Characterization of untethered robot: If not mentioned otherwise, untethered robots were operated in a tank of dimensions 130x50x50 cm³. The fill level of water was about 15 cm. A cluttered environment, simulating a rice field, was set up in a small inflaSupplementary Table wimming pool, using cat grass as plants. Except for the autonomous robot, robots were remote controlled using the aforementioned infrared controller. For the autonomous robot, we used various commercial white light sources, which were manually controlled. Swimming speed and trajectories were taken from image analysis using Kinovea software.

## Supporting Discussion

### Comparison of swimming robots.

In our comparison of swimming robots, we are using average speeds for linear motion and rotation. The speed is averaged over at least two actuation cycles. Motion tracking, using Kinovea software was used, if no values but videos were available in literature. We normalize the speed with respect to the robot's characteristic size, which is its largest dimension. This gives better comparison between robot designs with different aspect ratios.

### Actuator performance and power consumption.

Soft electrohydraulic actuators are capacitive transducers that operate under high electric fields. Several works have derived analytical models for this class of actuators(*31, 36, 37*). Here we highlight a few aspects that guide our discussion.

In their unactuated (relaxed) state a solid dielectric (thickness $t_s$, permittivity $\varepsilon_s$) and a liquid dielectric (thickness $t_{l,0}$, permittivity $\varepsilon_l$) are separating the electrodes. We note that $t_{l,0}$ is not constant but a function of the spatial coordinates $x$, and $y$. In the actuated (zipped) state, the electrodes are separated by the solid dielectric and a thin layer of liquid dielectric (thickness $t_{l,1}$). We approximate that the thickness of the liquid dielectric is constant over the zipped region. The capacitance ($C_1$) of the zipped region (area, $A$) is then given by

$$C_1 = (A\, \varepsilon_s\, \varepsilon_l)/(t_s\, \varepsilon_l + t_{l,1}\, \varepsilon_s)$$

We term this quantity the effective capacitance, as it is the governing term in calculating output force and power consumption of the actuator. We note that the capacitance of the unzipped region also contributes to both output force and power consumption, but its contribution can be neglected for typical actuator geometries.

Actuation of electrohydraulic actuators involves periodic charge and discharge of the capacitive structure with voltage ($U$); the frequency equals the actuation frequency ($f$). The power ($P_{in}$) that is consumed in this process is given by:

$$P_{in} = C_1\, f\, \frac{U^2}{2}$$

It is a linear function of actuation frequency and effective capacitance and a quadratic function of the applied voltage. When measuring the power consumption for a fixed voltage as a function of frequency, we obtain the effective capacitance through a linear regression of the measurement.

For a fixed voltage, the force of the actuator, given by the Maxwell pressure, is governed by the change of capacitance with respect to displaced volume of liquid dielectric ($\Delta V$). The displaced volume is the amount of liquid dielectric that is pushed by the zipping of the electrodes:

$$F_{el} \approx \frac{1}{2}\, U^2\, C_1{'}(\Delta V) \sim (G\, \varepsilon_s\, \varepsilon_l)/(t_s\, \varepsilon_l + t_{l,1}\, \varepsilon_s)$$

Where $G$ summarizes quantities that relate to the geometry of the actuator and can be a function of $\Delta V$. For a given actuator design, the force of the actuator depends mainly on the permittivities and thicknesses of the dielectric layers in the zipped state. Using a relative permittivity of 2.2 and thickness of 12 µm for PET, and 40 and 14 µm for the PVDF-terpolymer, the capacitance and force increase by a factor of 7.8. To achieve the same output force, the driving voltage can be reduced by a factor of 2.8, which matches well the obtained experimental results.

**Power consumption of the HVPS and pathway to energy autonomous operation.**

For the untethered robot, most power is consumed by the PCB including the voltage amplification by the HVPS. When the PCB is in idle state, where no voltage amplification occurs, it consumes 237 mW. Amplifying the voltage but having the actuators switched off increases the power consumption to 530 mW. When both actuators are driven with 30 Hz, the power consumption reaches 595 mW, indicating an additional 65 mW that is required to drive the actuators. Operation of the actuators makes only about 10% of the total power consumption. Strategies to achieve higher system efficiencies have to target the HVPS design. Solutions leading to highly efficient voltage amplification already exist(38), however are not commercially available yet, and are work-intensive to integrate in a customized PCB. Translating current academic efforts towards commercial products will immensely boost the potential of the introduced robot technology.

Currently, the 30 mAh battery is sufficient to drive the robot for about 10 min. Including a solar cell on top of the robot would be a strategy to increase the operational time of the robot and allow recharging of the battery. Monocrystalline Si-based solar cells, which have the same size as our robots, already provide sufficient power to recharge the battery. Optimization of the control circuit and HVPS will further reduce the consumed power drastically.

**Cantilever blocked force measurements**

To measure the blocked propulsion force ($F$) of the robot, the robot swims against a hollow glass cantilever. From the deformation of the cantilever, the blocked force is calculated. We used a hollow glass cantilever (wall thickness: 0.05 mm, width: 4 mm, thickness: 1 mm) that is clamped from the top. We determined the product of Young's modulus ($E$) and moment of inertia ($I$) through measuring the resonance frequency ($f$ = 55.576 Hz) of the cantilever, resulting in:

$$EI = \frac{4\pi^2}{3.5^2} f^2 \rho l^4 = 0.18775 \text{ m}^2 \text{ mN}$$

Where $\rho$ = 0.936 mg/mm is the specific mass per length and $l$ = 67 mm is the length of the (unclamped) cantilever. The deflection ($d_{meas}$) of the cantilever is measured at distance $a$ from the clamped part of the cantilever using the laser-displacement sensor. The robot is swimming against the cantilever at position $b$ = 66 mm (measured from the top). The deflection profile $d(x, F)$ at position $x$ of the cantilever (measured from the top) is then given by

$$d(x, F) = \frac{F x^2}{6 EI}(3b - x), for\ 0 < x < b$$

$$d(x, F) = \frac{F b^2}{6 EI}(3x - b), for\ b < x < l$$

We obtain the blocked propulsion force by solving:

$$d_{eff} = d(a, F)$$

### Wake structures during surface swimming

The swimming of the robot on the water surface creates a recirculation region behind the robot. The recirculation region is fed by two rearward moving jets generated on either side of the robot from its flapping motion during locomotion. The velocity of the rearward jets subtracted from the robot's speed provides the magnitude of these jets in the robot's frame of reference. The width of the wake generally results in smaller form drag on a moving body. A smaller wake width at 40 Hz compared to higher actuation frequencies suggests a possible explanation for the robot's optimal swimming performance at this frequency.

## Supplementary Figures

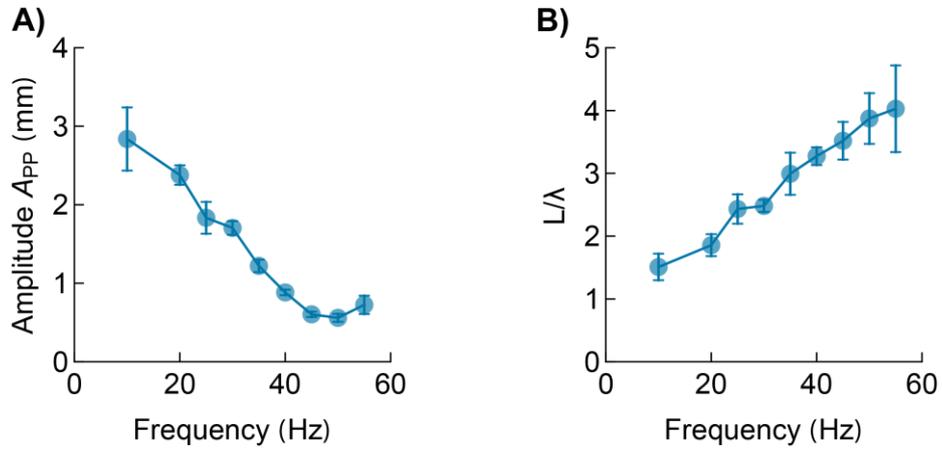

**Supplementary Fig. 1 | Dependence of wave metrics on the actuation frequency underwater. (A)** Peak-to-peak-amplitude as a function of actuation frequency at a driving voltage of 1500 V. **(B)** Number of waves present on the fin (length of fin $L$) as a function of frequency at a driving voltage of 1500 V. The increasing number of waves is equivalent to a decreasing wavelength ($\lambda$). Bipolar, rectangular signals were used for the driving voltage at an amplitude of 1500 V.

Manuscript Preprint: Highly agile flat swimming robot

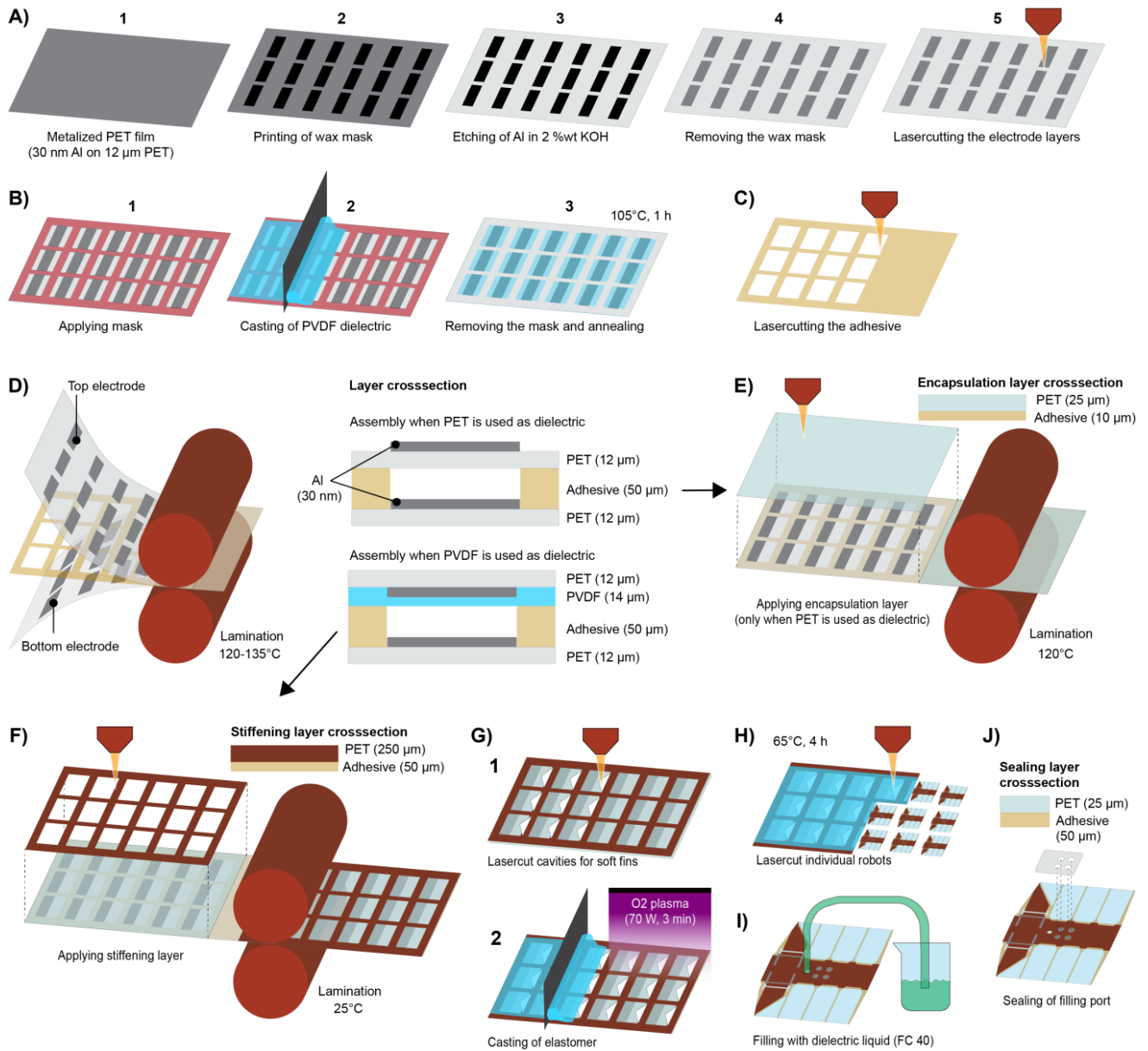

**Supplementary Fig. 2 | Fabrication of monolithic locomotion modules.** **(A)** Patterning of aluminum electrodes on 12-µm-thick PET substrates. Electrodes are patterned by etching, using printed wax as an etch-mask. **(B)** Patterning of PVDF-terpolymer dielectrics. Dielectric is cast through an acrylic mask and dried in an oven for 1 h at 105°C. **(C)** Structuring of thermal adhesive layer using laser cutting. **(D)** Lamination of top- and bottom-electrodes using the thermal adhesive. For layers with the PVDF-terpolymer dielectric, the dielectric faces inwards while the PET substrate serves as an encapsulation of the top electrode. When PET is used as dielectric, the substrate faces inwards and an additional encapsulation is required. **(E)** Application of 25-µm-thick encapsulation using a 10-µm-thick thermal adhesive. **(F)** Structuring and application of the stiffening layer, utilizing a 50-µm-thick acrylic adhesive. **(G)** Making the soft fins by, first, cutting the cavities for the soft fins. In a second step, the bottom side of the layer-structure is activated by O$_2$ plasma and a silicone elastomer is cast over the entire structure. **(H)** Individual locomotion modules are cut from the layer structure using laser cutting. **(I)** Filling of modules, using a single filling port and a pressure controlled reservoir of the dielectric fluid. **(J)** Sealing of filling port using an acrylic adhesive on PET.

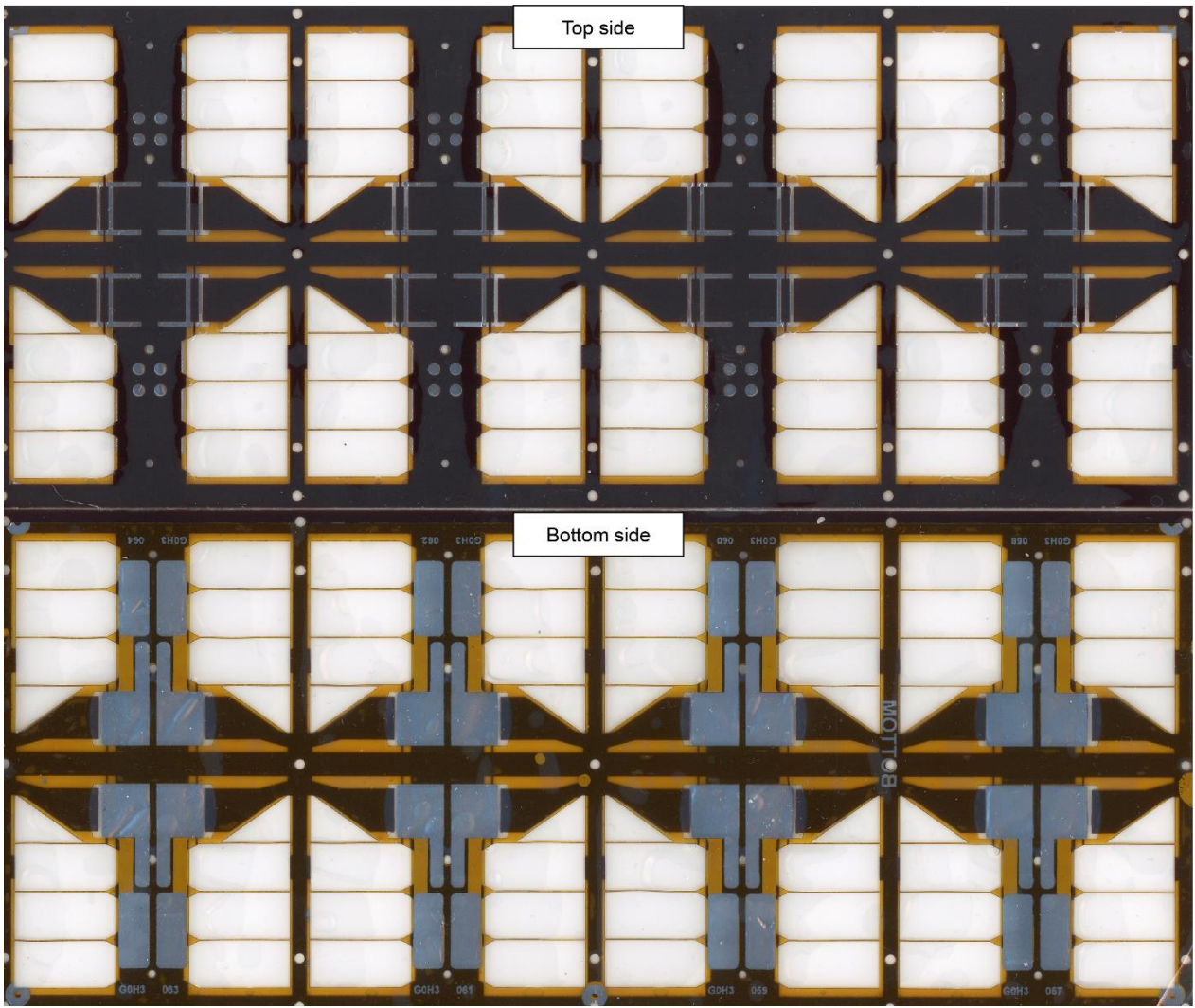

**Supplementary Fig. 3 | Grid of fabricated locomotion modules.** Top and bottom side of eight 2-actuator modules.

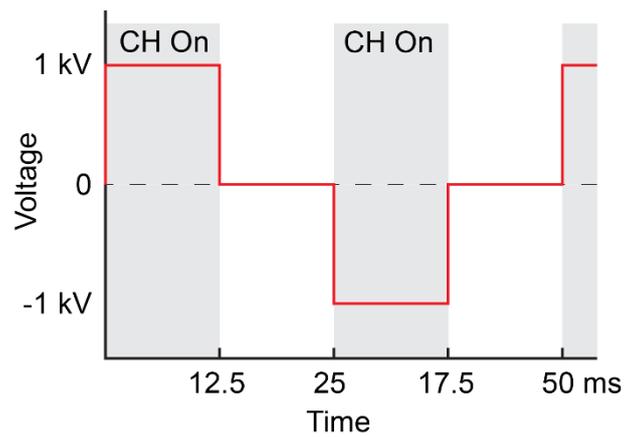

**Supplementary Fig. 4 | Schematic of driving signal for electrohydraulic actuators.** As electrohydraulic actuators respond to the square of the input voltage, the actuation frequency is twice the frequency of the bipolar driving signal.

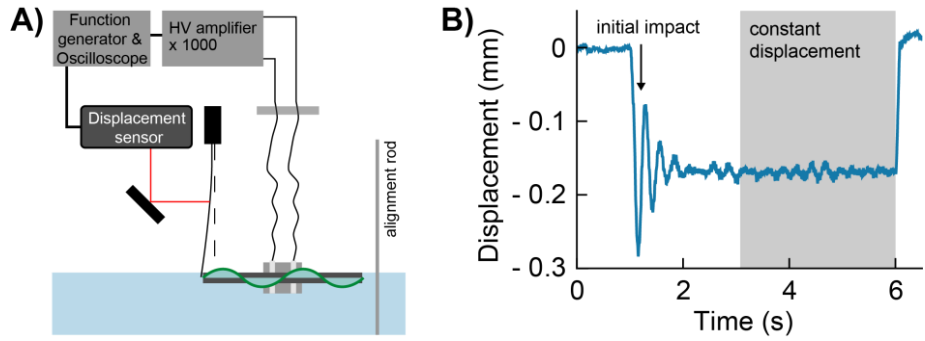

**Supplementary Fig. 5 | Blocked propulsion force measurement. (A)** Schematic of the setup and its involved components. **(B)** Measurement of the displacement of the cantilever as a function of time. After an initial impact the displacement stabilizes. The last 2.5 s of the measurement are taken to calculate the blocked propulsion force.

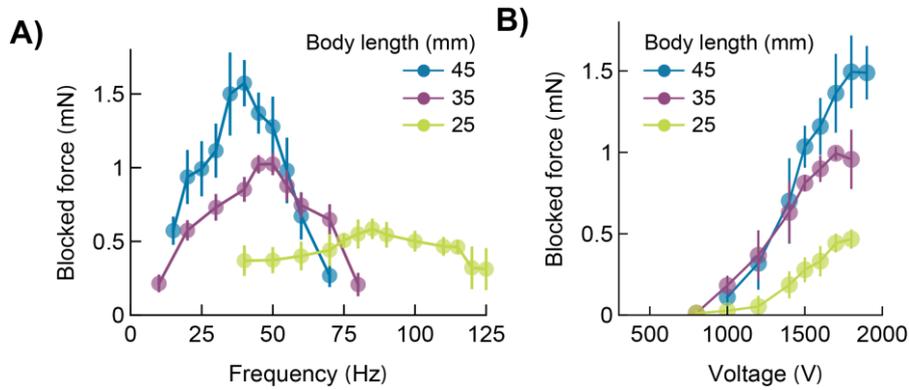

**Supplementary Fig. 6 | Detailed characteristics of propulsion modules with constant aspect ratio. (A)** Frequency dependence of three module sizes measured at 1700 V. **(B)** Voltage dependence for three module sizes, measured at their optimal actuation frequency.

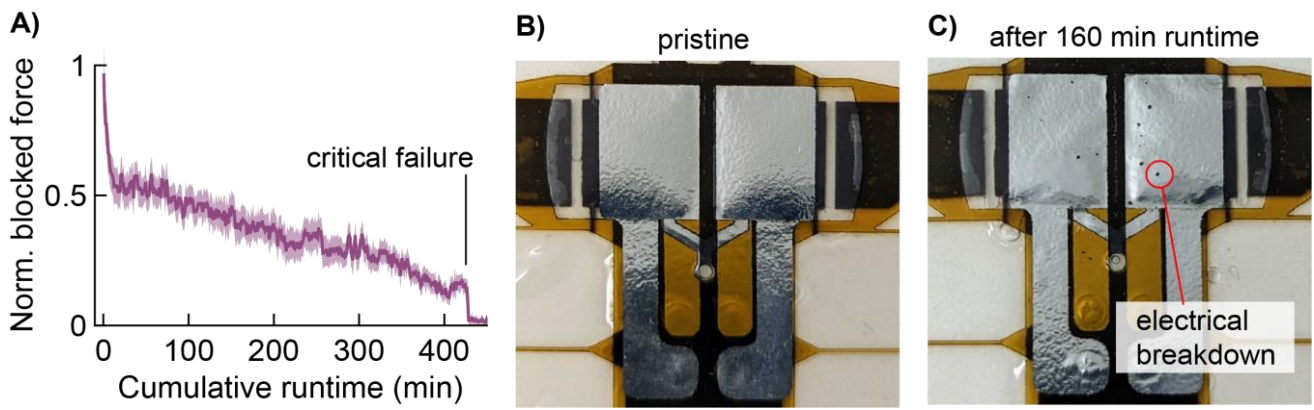

**Supplementary Fig. 7 | Lifetime and fatigue of propulsion modules. (A)** Normalized blocked force as a function of cumulative runtime, measured at 30 Hz actuation frequency. The module uses the PVDF-terpolymer as solid dielectric and achieved 427 min runtime, equivalent to 768,600 actuation cycles, before critical failure. The module was on the water surface for a total time of 42.7 h. **(B)** Image of a pristine propulsion module (bottom view) using PET as dielectric. **(C)** Image of a propulsion module, using PET as dielectric, after 160 min of cumulative operation at 40 Hz actuation frequency (equivalent to 380k actuation cycles). Multiple electrical breakdown events occurred, yet, did not interfere with operation due to local ablation of the thin aluminum electrodes. Through the repeated operation, fatigue in actuator mechanics becomes visible. The liquid-filled pouches are losing tension.

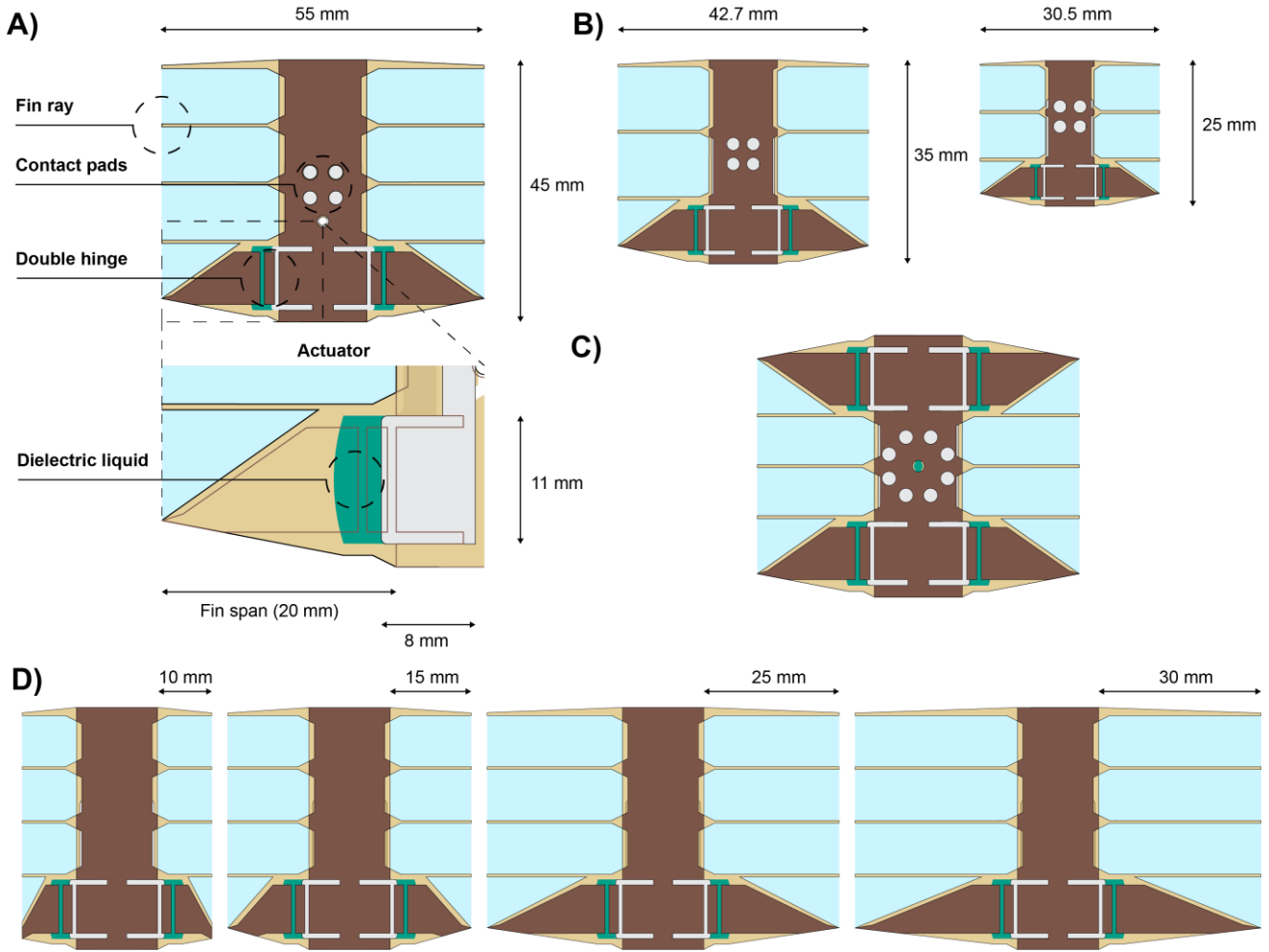

**Supplementary Fig. 8 | Designs of flat locomotion modules. (A)** 2-actuator design with 45 mm length and 55 mm width. The stiffening layer features a double-hinge and fin rays. The fin rays are used to allow better adhesion of the silicone membrane to the main body. Two cutouts in the stiffening layer serve to amplify the bending amplitude. **(B)** Scaled designs, which maintain the same aspect ratio as in (A). The smallest module fabricated has a length of 25 mm. **(C)** 4-actuator design for enhanced locomotion capabilities; it has a length of 45 mm and a width of 55 mm. **(D)** Designs with scaled fin width. The size of actuators and body length is the same as within the design in (A).

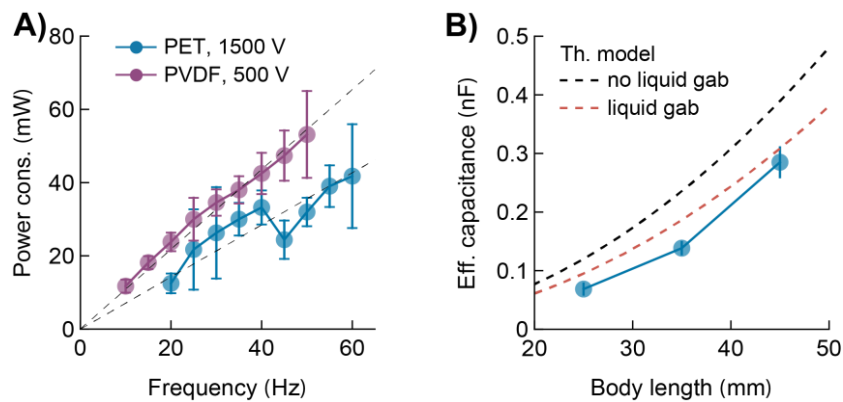

**Supplementary Fig. 9 | Power consumption of tethered propulsion modules. (A)** Frequency dependent power consumption of modules using PET or PVDF-terpolymer as solid dielectric. Dashed lines, linear fit using the law for the power consumption when charging capacitors. **(B)** From the linear fit, the effective capacitance, the highest capacitance in the actuators zipped state, is obtained. Dashed lines represent models for the actuators' capacitance under uniform areal scaling. Red dashed line, represents the capacitance assuming a liquid gap of 2 μm (see Supplementary Discussion).

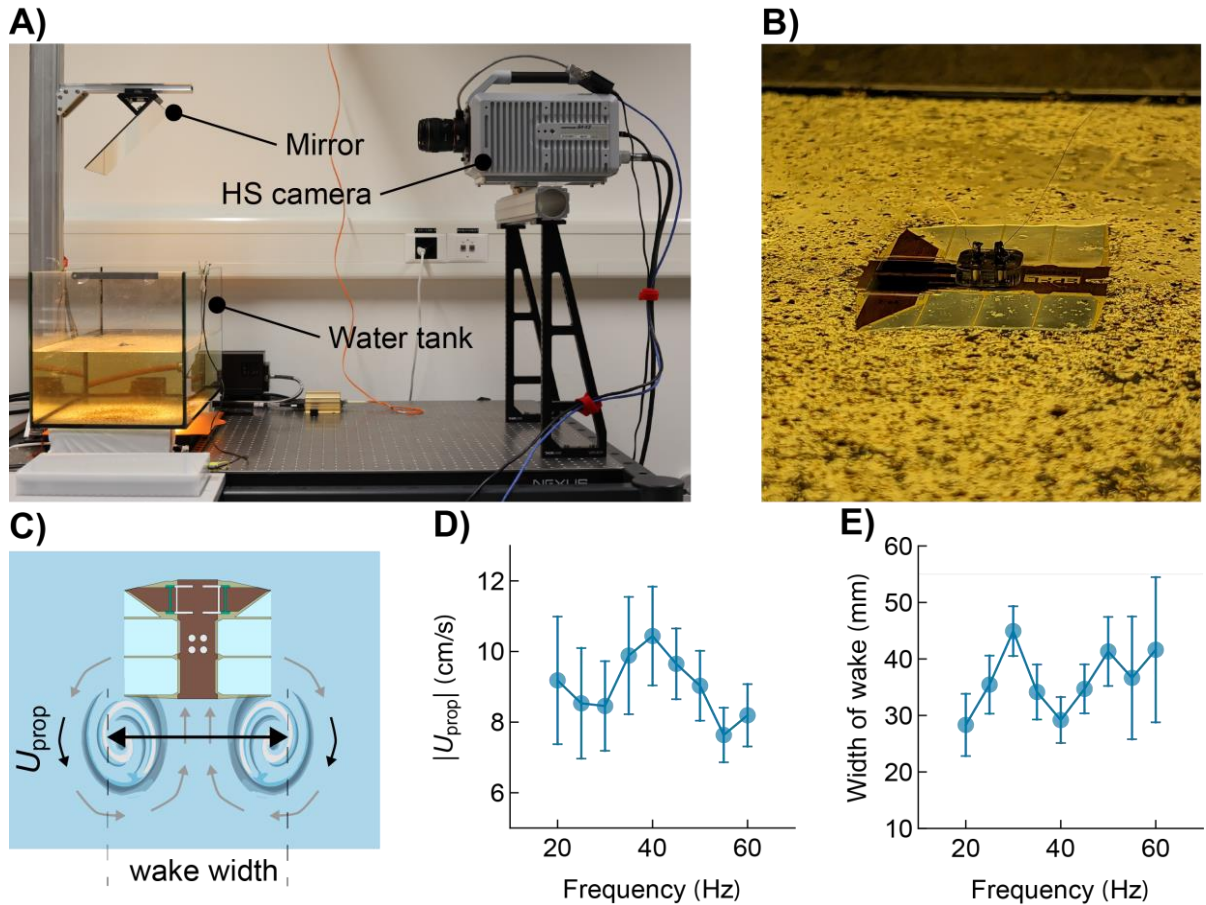

**Supplementary Fig. 10 | LIC measurement setup and wake analysis. (A)** Setup to image water surface flows. A high-speed camera is capturing the water surface through a mirror mounted at 45 degrees. **(B)** Tethered locomotion module swimming on the water surface. Patterns of milled pepper particles are cross-correlated to analyze the flow. **(C)** Schematic of the main metrics characterizing the flow in the wake of the robot. We characterize the rearward propulsion speed ($U_{prop}$) of water streams and the width of the wake behind the robot. **(D)** $U_{prop}$ as a function of actuation frequency shows a maximum of 40 Hz, correlating to the frequency at which we observe maximum swimming speed of the robot. **(E)** Width of the wake as a function of frequency, is minimal at the optimal actuation frequency.

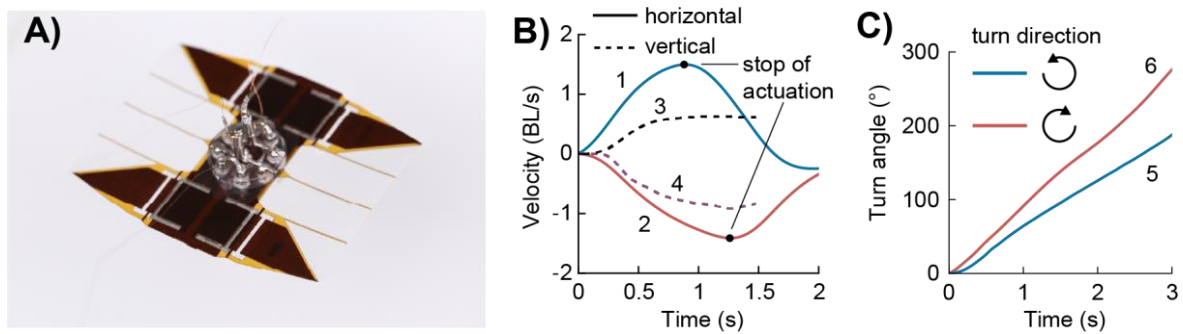

**Supplementary Fig. 11 | Linear and rotation speed of 4-actuator module. (A)** Image of 4-actuator module floating on the water surface (BL, 45 mm). **(B)** Forward, backward, and sideways swimming speeds as a function of time. The actuation of the robot was stopped by the operator, after a few body lengths of translation. **(C)** Rotation angle versus time for clockwise and counterclockwise rotation.

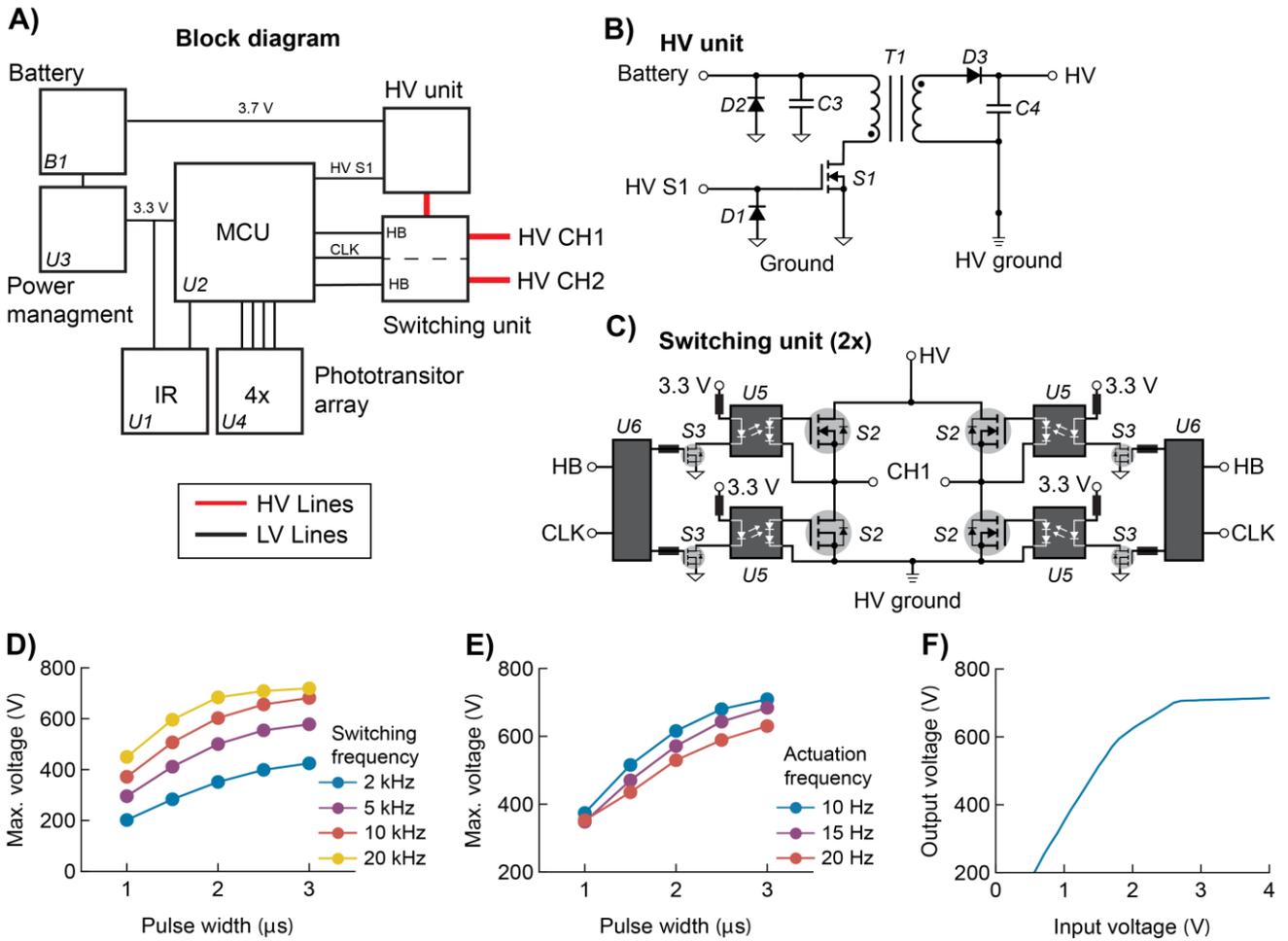

**Supplementary Fig. 12 | Detailed schematics of the driving electronics and its characteristics. (A)** Block diagram of circuit board components and its connections. Red lines mark HV lines. **(B)** Schematic of the flyback converter (FBC) circuit. **(C)** Schematic of the switching unit, consisting of two full bridges, supplying bipolar HV signals to both actuators. **(D)** Maximal output voltage of the FBC as function of pulse width and for various switching frequencies of the transistor $S_1$. There is no load (actuators) connected to the output of the FBC. **(E)** Maximal output voltage of the FBC under load and as a function of pulse width for three different actuation frequencies. The switching frequency of transistor $S_1$ is set to 20 kHz in each case. **(F)** FBC output voltage as a function of input voltage.

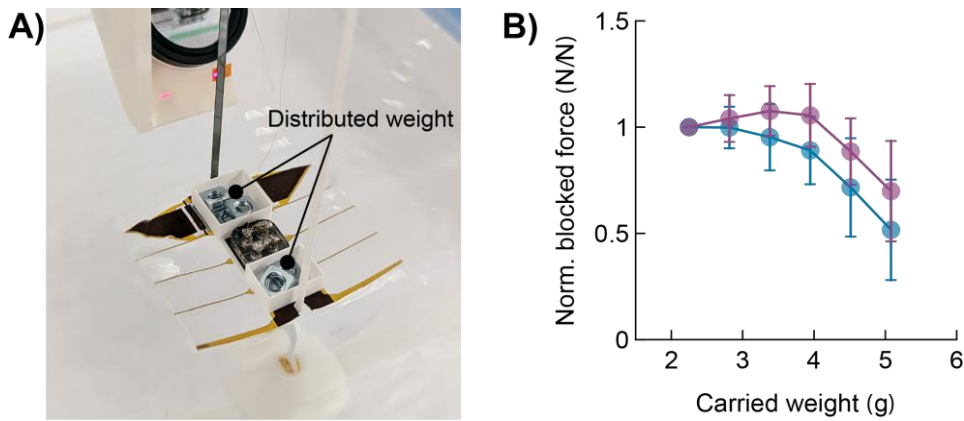

**Supplementary Fig. 13 | Swimming performance under additional load. (A)** Image of a locomotion module with distributed weights on the front and the rear part of the module (BL, 45 mm). **(B)** Normalized block force as a function of added weight. Loading the robot with more than 5.1 g, would result in the module to sink. Blue, 1500 V. Purple, 1700 V.

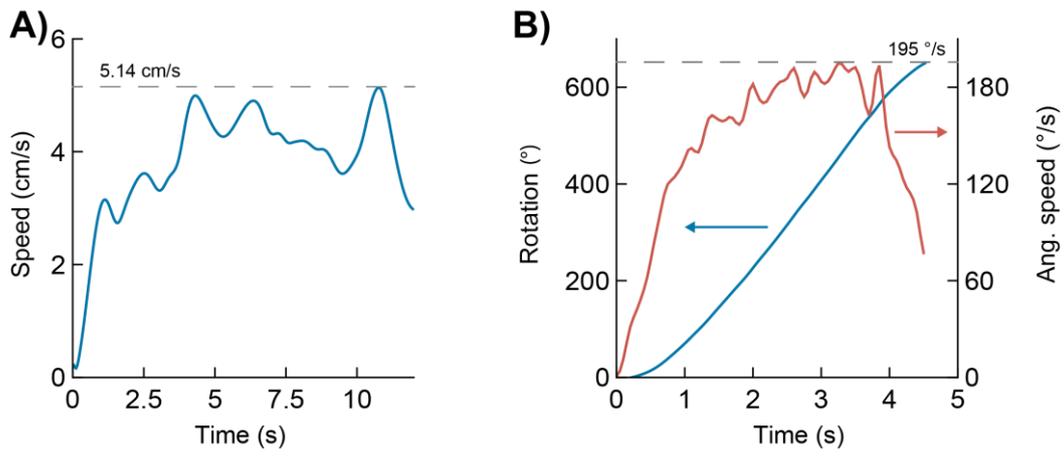

**Supplementary Fig. 14 | Swimming and rotation speed for untethered robots.** **(A)** Linear swimming speed for an untethered autonomous robot following a light source. **(B)** Rotation angle and angular speed of an untethered remote controlled robot.

## Supplementary Tables

| | Locomotion type | Actuator Technology | Tethered / Untethered | Surface / Underwater | L (mm) | W (mm) | H (mm) | Weight (g) | Swimming Speed (cm/s) | (BL/s) | (CS/s) | Maneuverable / Rotation Speed (deg/s) |
|---|---|---|---|---|---|---|---|---|---|---|---|---|
| 1 | Oscillation | Magnetic | Tethered* | S | 9 | 10 | 4.7 | 0.18 | 4.3 | 4.8 | 4.3 | |
| | | | Untethered* | S | 20 | 22 | 15 | 4.73 | 3.6 | 1.8 | 1.64 | |
| 2 | Vibration | Piezoelectric | Tethered | S | 40 | 74 | 23 | 0.88 | 1.85 | 0.46 | 0.25 | ✓ 6 |
| 3 | Vibration | DC (vibration) motor | Untethered | S | 50 | 30 | 11 | 2.6 | 1.8 | 0.36 | 0.36 | |
| 4 | Vibration | DC (vibration) motor | Untethered | S | 75 | 95 | 21 | 35 | 17.1 | 2.28 | 1.8 | ✓ 142 |
| 5 | Pectoral fin (Type C) | DC motor | Tethered | S | 95 | 155 | N/A | 22.7 | 11.7 | 1.23 | 0.75 | |
| 6 | Pectoral fin (Type C) | DEA | Tethered | S/(U) | 100 | 123 | 66 | 11.5 | 7.6 | 0.76 | 0.62 | ✓ 18 |
| 7 | Pectoral fin (Type C) | Electroosmotic hydrogel | Untethered | S | 43 | 15 | 8 | 1.45 | 0.71 | 0.16 | 0.16 | |
| | | | Untethered | S | 35 | 47 | 10 | 2.51 | 0.78 | 0.22 | 0.17 | ✓ 15.4 |
| 8 | Pectoral fin (Type C) | Piezoelectric | Tethered | S | 40 | 20 | 20 | 1.65 | 2.8 | 0.7 | 0.7 | ✓ 27.7 |
| 9 | Water strider | Piezoelectric | Tethered | S | 100 | 100 | N/A | 1 | 3 | 0.3 | 0.3 | ✓ 28.6 |
| 10 | Water strider | DC motor | Untethered | S | 150 | 150 | 8 | 6.1 | 8.7 | 0.58 | 0.58 | (✓) 45.8 |
| 11 | Maragngoni effect | - | Untethered | S | 72.5 | 65.5 | 10 | N/A | 2 | 0.28 | 0.28 | |
| 12 | Maragngoni effect | - | Untethered | S | 52.4 | 32.3 | N/A | N/A | 2.5 | 0.48 | 0.48 | |
| 13 | Jelly fish | Electrohydraulic actuator (HASEL) | Tethered | U | 160 | 160 | N/A | 50 | 6.1 | 0.38 | 0.38 | ✓ - |
| | | | Untethered | U | 160 | 160 | N/A | 170 | 2 | 0.13 | 0.13 | |
| 14 | Jelly fish | Hydrogel | Untethered* | U | 10 | 10 | N/A | N/A | 0.33 | 0.33 | 0.33 | |
| 15 | Jet propulsion | DEA | Tethered | S | 76 | 50 | 50 | 54 | 5 | 0.66 | 0.66 | |
| | | | Tethered | U | 76 | 50 | 50 | 54 | 3.3 | 0.43 | 0.43 | |
| 16 | Jet propulsion | DEA | Untethered | S/(U) | 55 | 95 | 95 | 126 | 2.1 | 0.38 | 0.22 | |
| 17 | Caudal fin | DC pump (hydraulic) | Untethered | U | 470 | 230 | 180 | 1600 | 23.5 | 0.5 | 0.5 | ✓ 14 |
| 18 | Caudal fin | DEA | Untethered | U | 100 | 60 | 30 | 115 | 5.5 | 0.55 | 0.55 | |
| 19 | Caudal fin | Biohybrid | Untethered* | S/(U) | 14 | N/A | N/A | 0.025 | 1.5 | 1.07 | 1.07 | |
| 20 | Pectoral fin (Type B) | Biohybrid | Untethered* | S/(U) | 16.3 | N/A | N/A | 0.01 | 0.32 | 0.2 | 0.2 | |
| 21 | Pectoral fin (Type B) | DEA | Untethered | U | 115 | 280 | N/A | N/A | 3.89 | 0.34 | 0.14 | |
| 22 | Pectoral fin (Type B) | DEA | Tethered | S/(U) | 93 | 220 | 40 | 42.5 | 13.5 | 1.45 | 0.61 | ✓ N/A |
| | | | Untethered | S/(U) | 93 | 220 | 40 | 90.3 | 6.4 | 0.69 | 0.29 | ✓ 3.8 |
| 23 | Pectoral fin (Type B) | Pneumatic (soft) | Tethered | U | 65 | 150 | 6.5 | 2.8 | 9.4 | 1.45 | 0.63 | (✓) 18.5 |
| 24 | Pectoral fin (Type B) | Rolled DEA | Tethered | S/U | 46 | 60 | 4 | N/A | 6.4 | 1.36 | 1.07 | ✓ N/A |
| 25 | Pectoral fin (Type A) | DC motor | Untethered | U | 370 | 190 | 50 | 430 | 18.5 | 0.5 | 0.5 | ✓ 30 |
| 26 | Pectoral fin (Type A) | DC motor | Tethered | U | 220 | 95 | N/A | N/A | 25 | 1.14 | 1.14 | - |
| 27 | Pectoral fin (Type A) | DC motor | Untethered | U | 407 | 340 | 64 | 2800 | 31.6 | 0.78 | 0.78 | ✓ 52.1 |
| 28 | Pectoral fin (Type A) | DC motor | Untethered | U | 240 | 570 | 100 | 2850 | 23 | 0.96 | 0.4 | ✓ 22 |
| | Pectoral fin (Type A) | Electrohydraulic actuator (HASEL) | Tethered | S | 45 | 55 | 0.5 | 1.25 | 11.9 | 2.64 | 2.16 | ✓ 120 |
| | | | Untethered | S | 45 | 55 | 3 | 6.25 | 5.14 | 1.14 | 0.93 | ✓ 195 |

**Supplementary Table 1 | Comparison of swimming robots. Locomotion type,** sorted into 10 main categories, types of pectoral fins are as depicted in Fig. 1. **Actuator technology,** soft or soft-hard hybrid actuation mechanisms are marked bold. **Tethered/Untethered,** untethered robots that are marked with a star operate only in special fluids, or require power or magnetic fields from external stationary infrastructure. **Surface/Underwater,** Robots that swim on or close to the water surface are labeled S, Robots that are designed for underwater operation but swim close to the water surface are labeled S/(U), Robots that operate underwater are labeled U. **Dimensions,** length (L), width (W), height (H), the largest dimension is the characteristic size (CS) and marked bold. **Swimming speed,** given in absolute and relative metrics. **Maneuverable/Rotation speed,** robots that are capable of controlled maneuvers are labeled with "✓", Robots that can not continuously switch between swimming forward or rotation are labeled with "(✓)". The data of this work is boxed. Index corresponds to reference, 1:(18), 2:(12), 3:(19), 4:(20), 5:(17), 6:(16), 7:(15), 8:(9), 9:(10), 10:(11), 11:(13), 12:(14), 13:(4), 14:(39), 15:(40), 16:(41), 17:(2), 18:(42), 19:(25), 20:(24), 21:(3), 22:(21), 23:(22), 24:(23), 25:(5), 26:(43), 27:(44), 28:(6).

| Designator | Value | Name | Description |
| --- | --- | --- | --- |
| B1 | 30 mAh | | Li-Po Battery |
| C3 | 22 µF | GRM158R60J226ME01D | Capacitor |
| C4 | 2 nF | C0805C202KDRAC7800 | Capacitor |
| D1,D2 | | BZX884-C3V6315 | Zehner Diode |
| D3 | | RFU02VSM8STR | Diode |
| R1 | 180 Ohm | | Resistor |
| S1 | | EPC2035 | Enhancement Mode Power Transistor |
| S2 | | BSS127S-7 | MOSFET N-CH 600 V |
| S3 | | NTNS3193NZT5G | Single N-Channel Small Signal MOSFET |
| T1 | | ATB322515-0110 | Transformer |
| U1 | | TSOP37238TT1 | IR Remote Receiver 38kHz |
| U2 | | MKL02Z32CAF4R | Microcontroller |
| U3 | | AEM10941-QFN | Power Management Chip |
| U4 | | TEMT6000X01 | Ambient Light Sensor |
| U5 | | APV2111VY | Optocoupler |
| U6 | | SN74LVC2G74YZPR | D-Type Flip-Flop |

**Supplementary Table 2 | List of main PCB components.**

## Supplementary Movie Captions

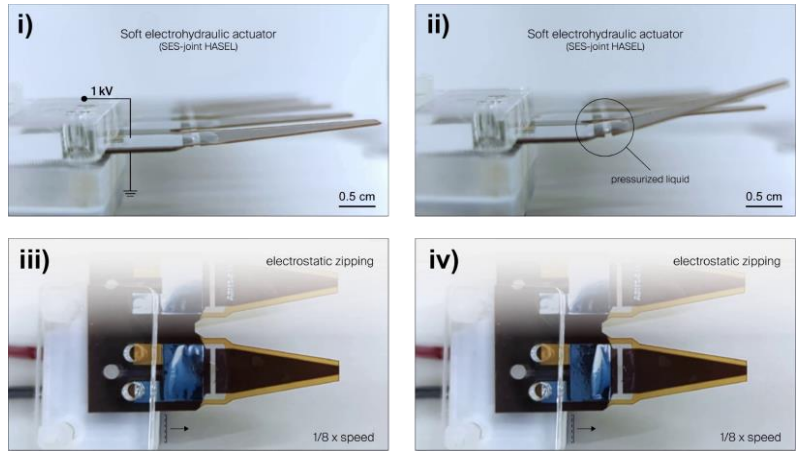

**Supplementary Movie 1 | Bending motion of actuators.** Electrohydraulic actuators based on the SES-joint HASEL design operating in air at 1 kV. Slow motion sequence showing the zipping of electrodes, which is induced by coulombic forces.

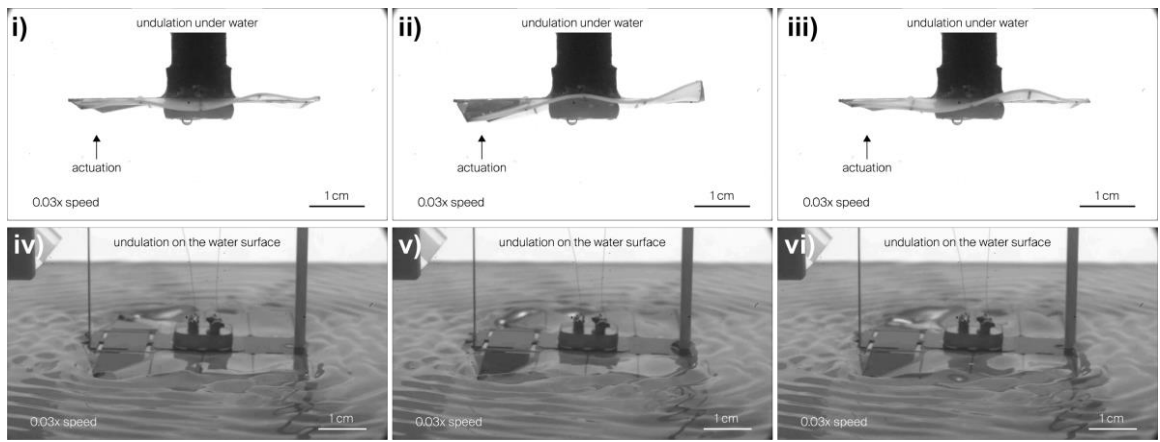

**Supplementary Movie 2 | Traveling wave generation in undulating fins.** A locomotion module is placed underwater and connected through a stiff rod to a HV power supply. A bipolar triangular wave is supplied with an amplitude of 2 kV and at 16 Hz actuation frequency.

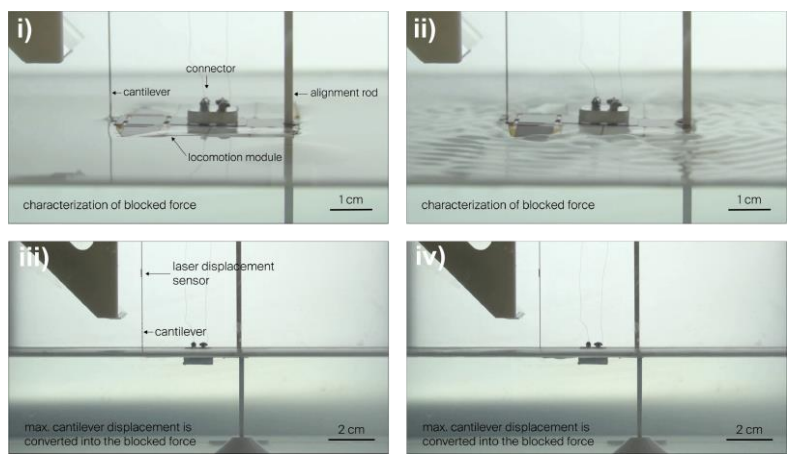

**Supplementary Movie 3 | A setup for blocked force measurements.** Tethered locomotion module swimming against a glass cantilever. The deformation of the cantilever is used to calculate the blocked propulsion force of the robot.

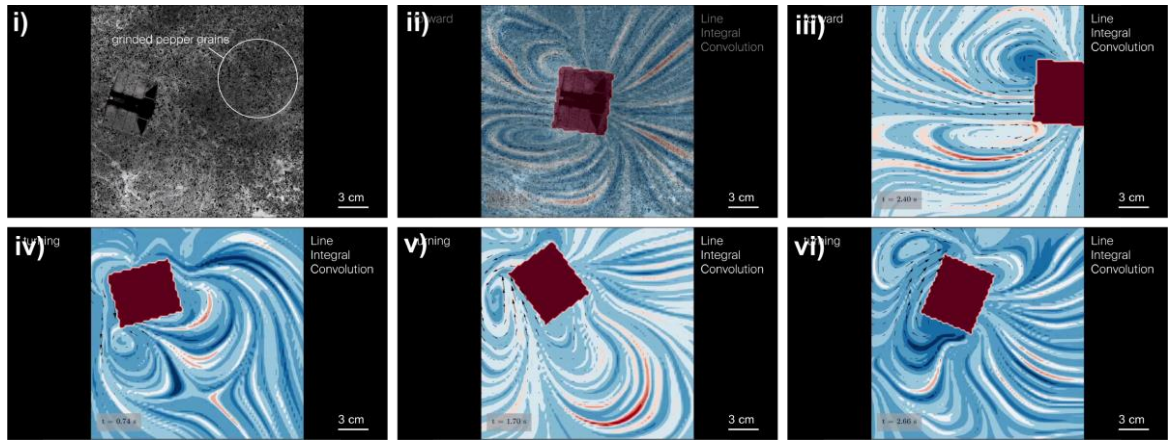

**Supplementary Movie 4 | Line integral convolution for flow visualization.** Tethered locomotion module swimming through milled pepper particles. Cross-correlation of particle movement results in a surface flow field. Streak lines are visualized using line integral convolution.

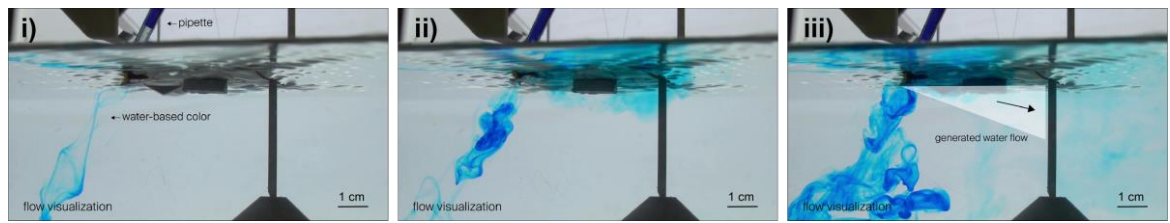

**Supplementary Movie 5 | Flow visualization under the water surface.** Drops of food color are dropped into the water in front of the robot. The propagation of the color is filmed using a high-speed camera.

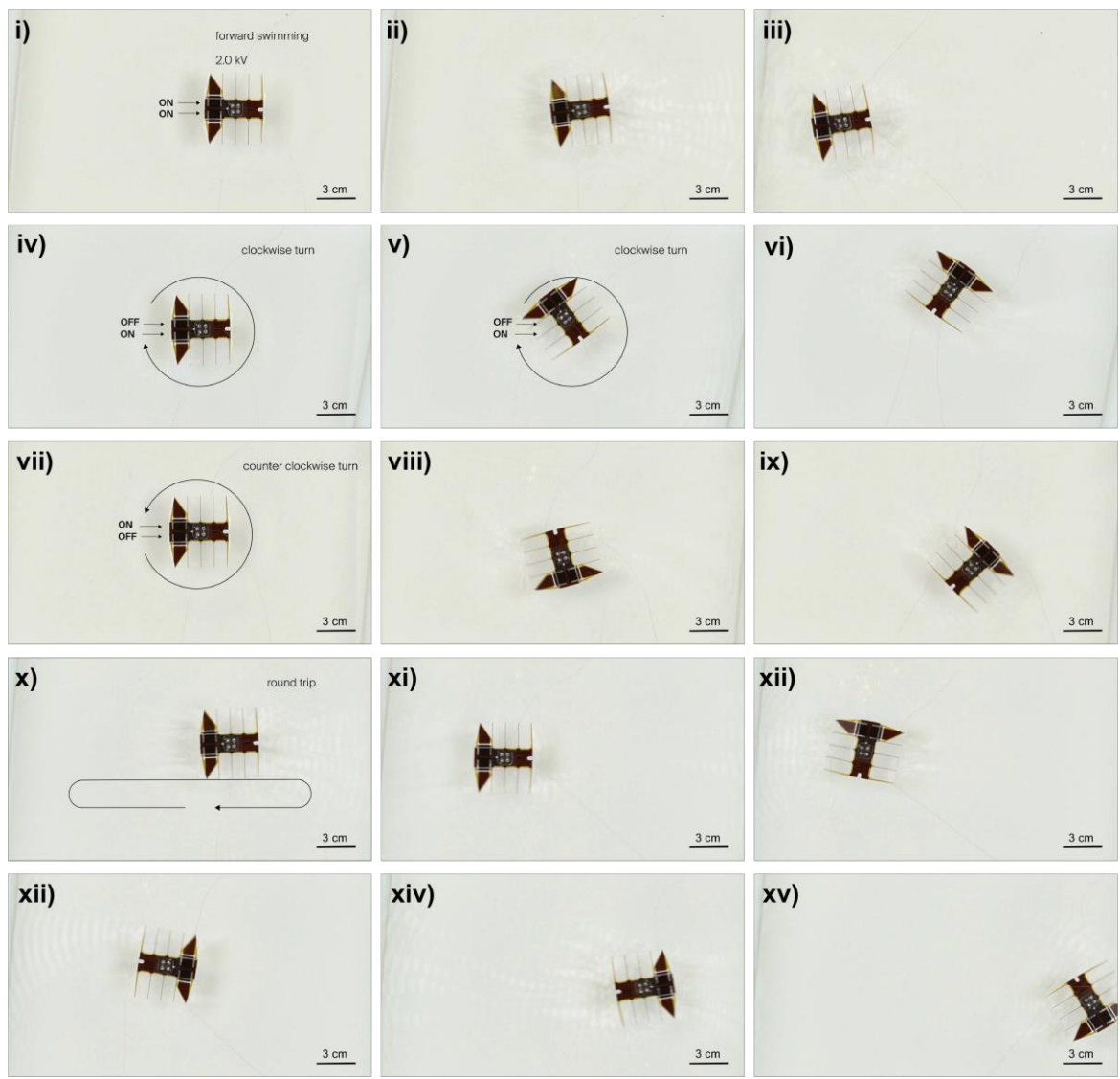

**Supplementary Movie 6 | Maneuverability of a 2-actuator module.** Tethered locomotion module in the 2–actuator design is connected to a HV power supply. Forward propulsion at 2 kV is shown, as well as rotation and mixed locomotion patterns (at 1.7 kV).

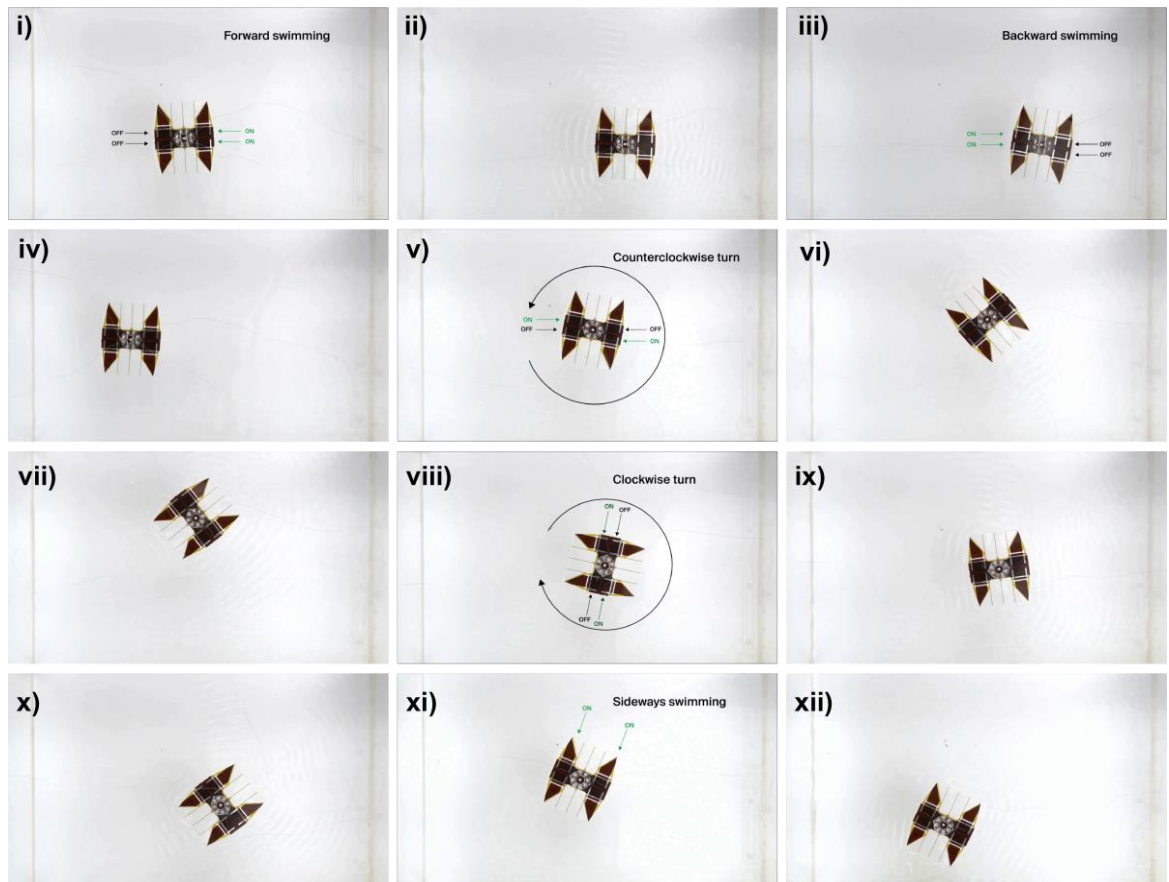

**Supplementary Movie 7 | Maneuverability of a 4-actuator module.** Tethered locomotion module in the 4-actuator design is connected to a HV power supply. Forward-, backward-, sideways propulsion, and rotation at a supplied voltage of 1.7 kV are shown.

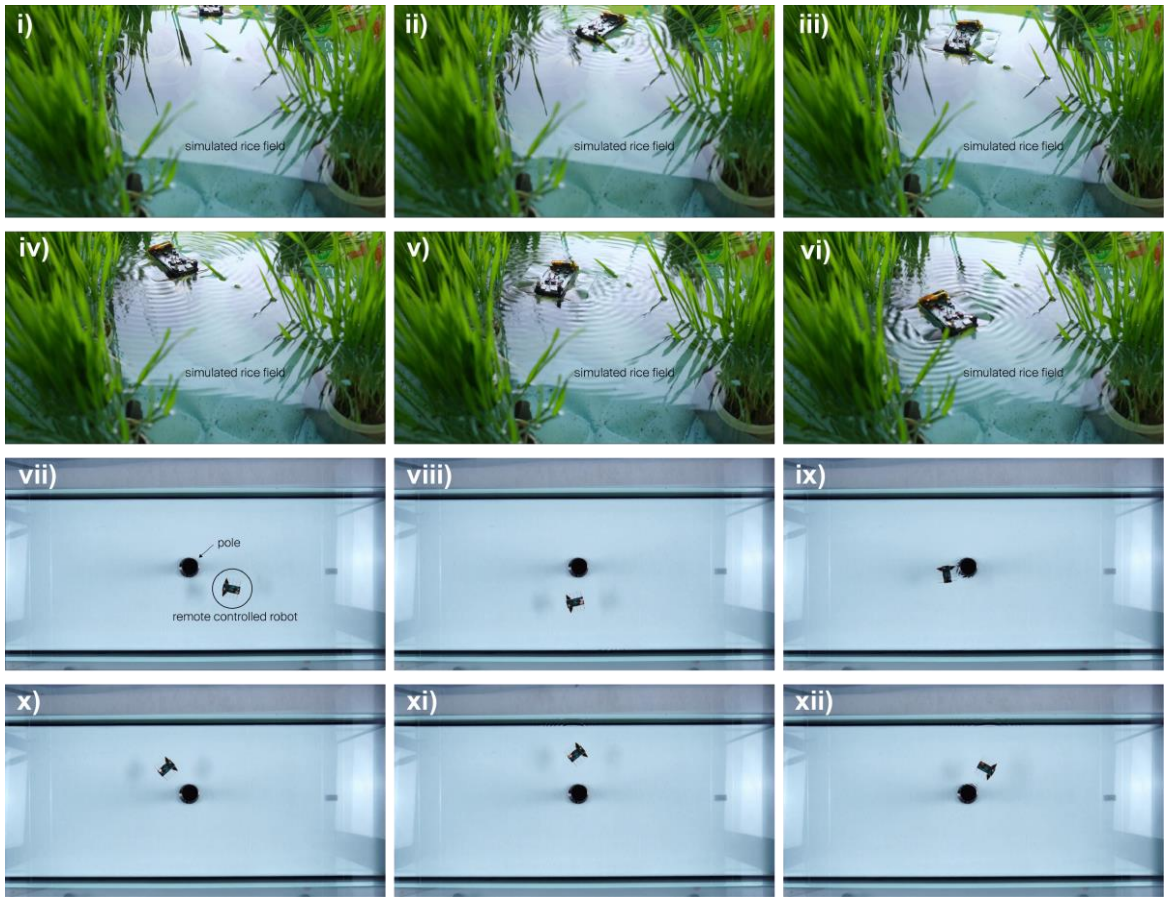

**Supplementary Movie 8 | Steering of an untethered robot.** An untethered, remote controlled robot is steered through a cluttered, grassy environment, and around a pole. The operator is using a 4-button remote control sending commands. Each press of a button results in propulsion with preprogrammed duration.

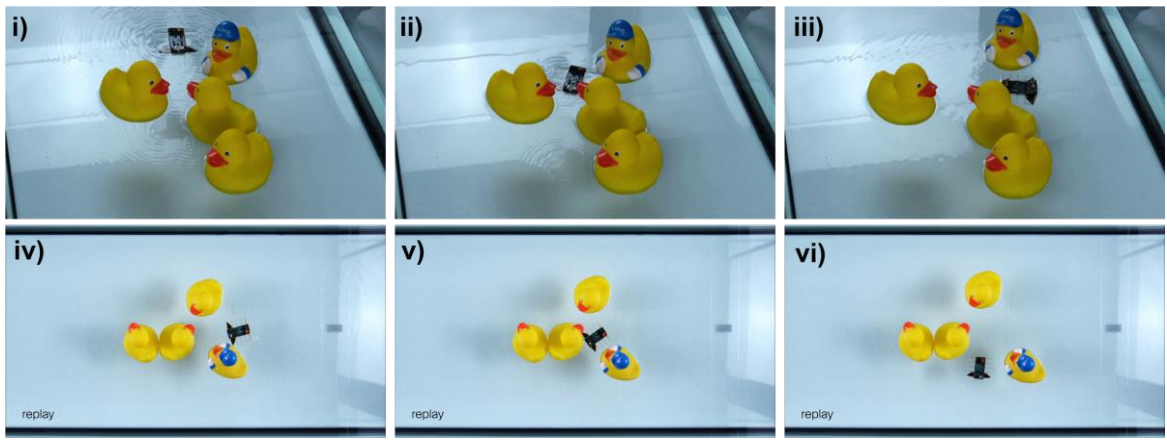

**Supplementary Movie 9 | Swimming through floating objects.** An untethered, remote controlled robot is steered through a set of floating objects. The gaps between objects are narrow, measuring about the size of the robot.

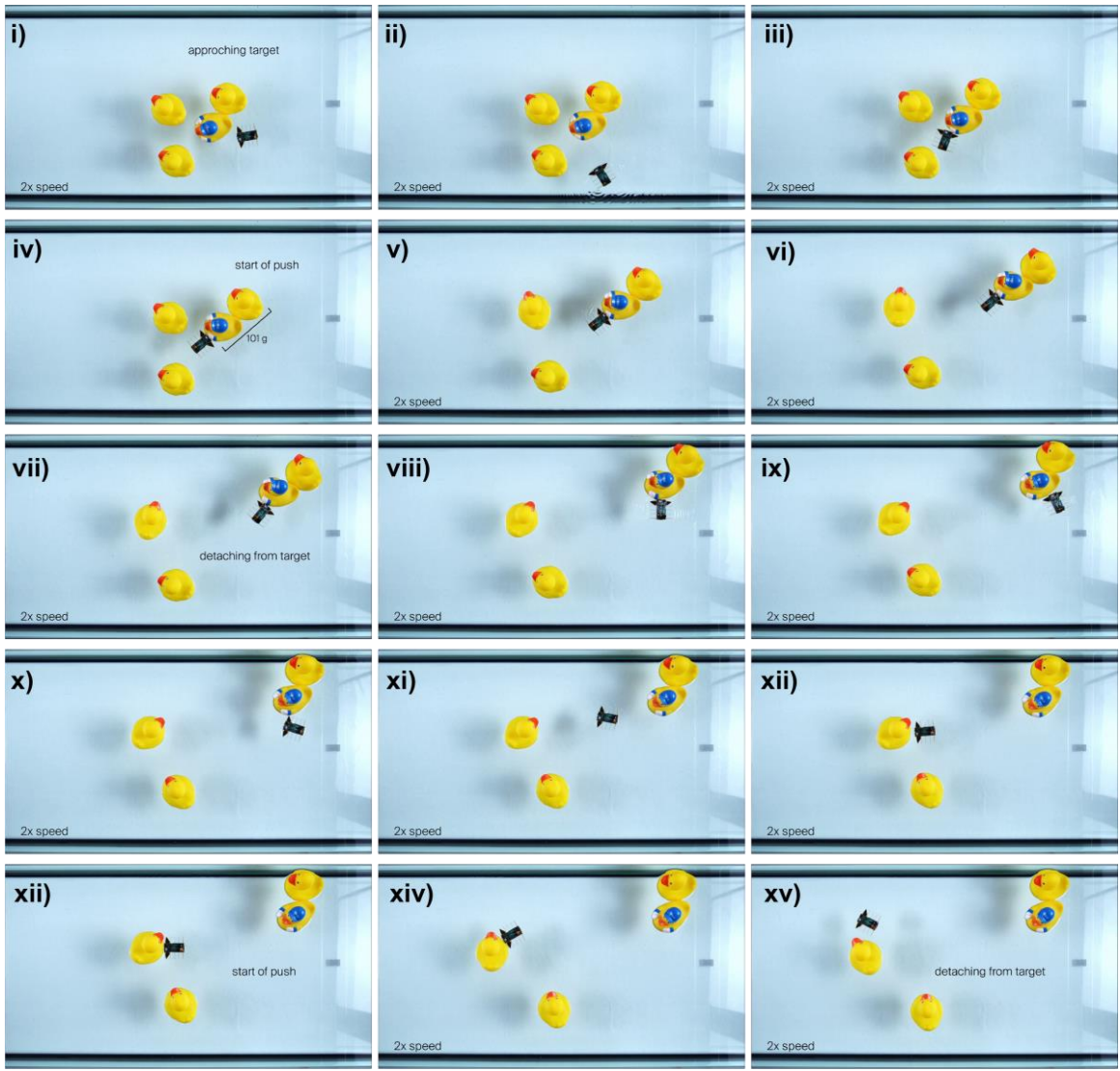

**Supplementary Movie 10 | Clearing of floating objects.** An untethered, remote controlled robot performs displacement of heavy floating objects. First, the robot is approaching the target objects which are 16x the weight of the robot. Second, it is displacing those objects for a distance of several body lengths. Third, the robot is detaching from the target objects through a series of sideways movements (initiated through rotation commands). Finally, the robot approaches the next target object and performs the same steps.

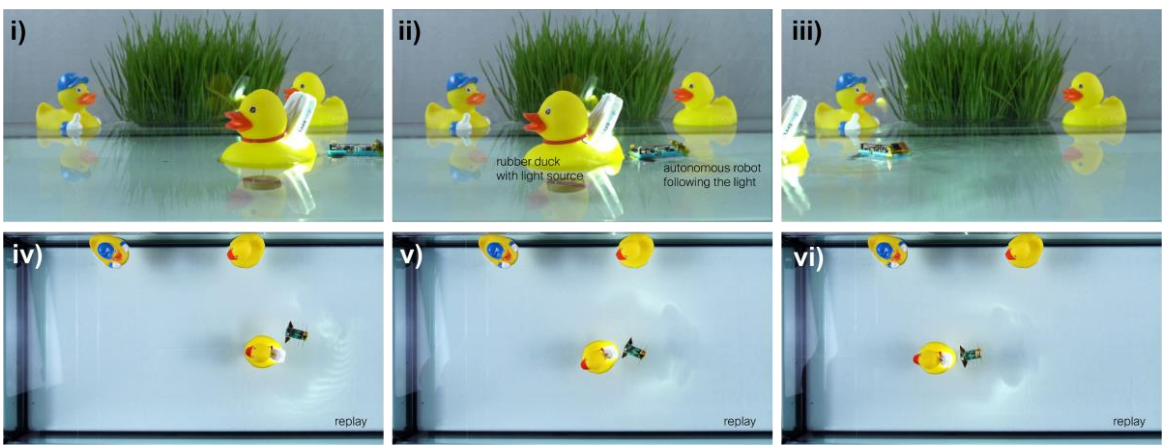

**Supplementary Movie 11 | Autonomous swimming, following a lead object.** An untethered, autonomous robot follows a lead object carrying a light source. The lead object is manually pulled using a nylon thread. The robot detects the direction of the light source and swims after the light source.

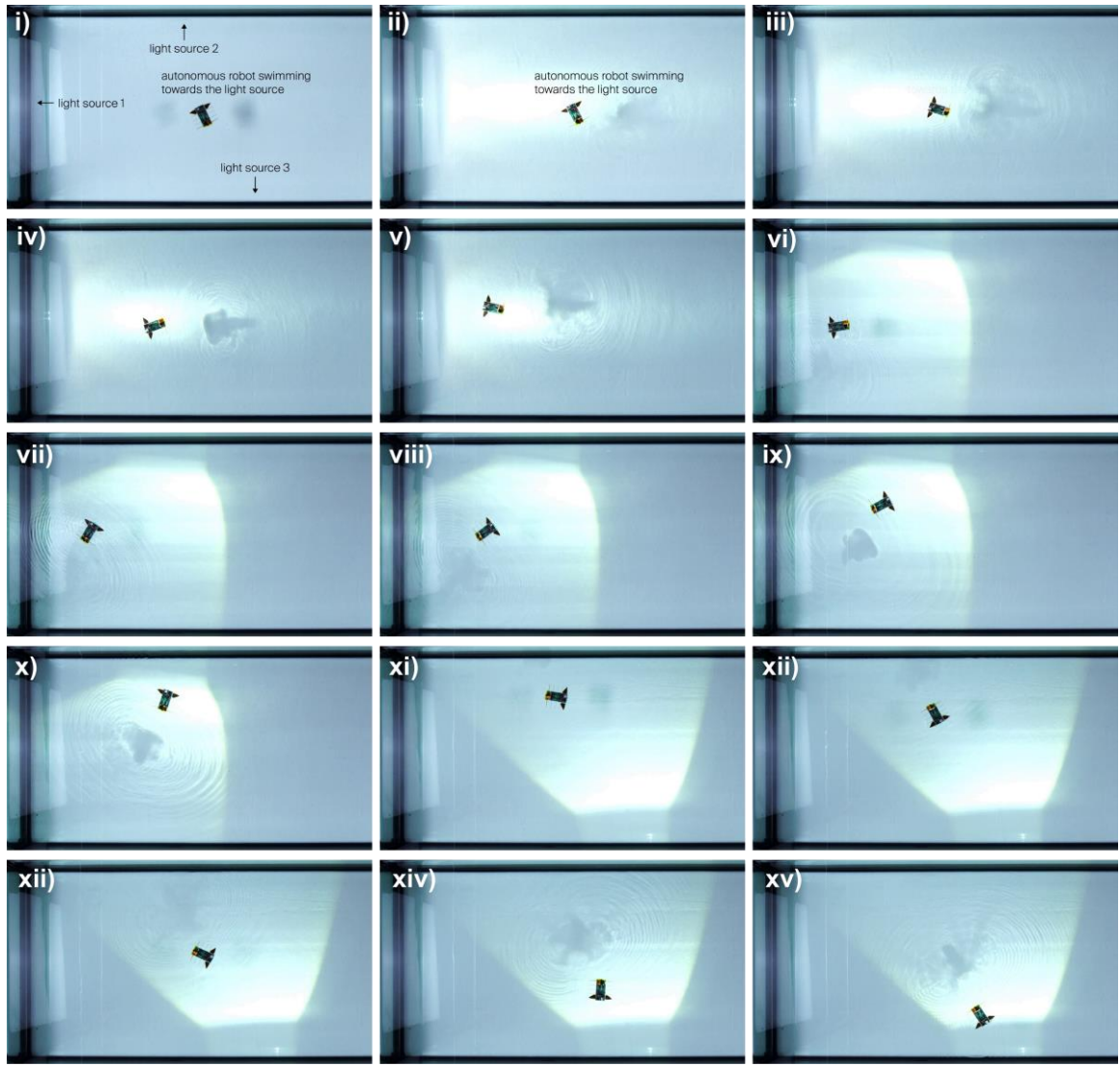

**Supplementary Movie 12 | Autonomous swimming, navigating towards lights.** An untethered, autonomous robot is homing towards a light source. Three light sources are activated manually in sequence. The robot detects the direction of the active light source and swims towards it.

## Supplementary References